\documentclass{article}

\usepackage[preprint]{neurips_2026}

\usepackage[utf8]{inputenc} 
\usepackage[T1]{fontenc}    
\usepackage{hyperref}       
\usepackage{url}            
\usepackage{booktabs}       
\usepackage{amsfonts}       
\usepackage{nicefrac}       
\usepackage{microtype}      
\usepackage{xcolor}         
\usepackage{siunitx}

\usepackage[utf8]{inputenc} 
\usepackage[T1]{fontenc}    
\usepackage{hyperref}       
\usepackage{url}            
\usepackage{booktabs}       
\usepackage{amsfonts}       
\usepackage{nicefrac}       
\usepackage{microtype}      
\usepackage{xcolor}         
\usepackage{placeins}

\usepackage{wrapfig}

\usepackage{makecell}
\usepackage{rotating}
\usepackage{enumitem}

\usepackage{times}
\usepackage{latexsym}

\usepackage[utf8]{inputenc}

\usepackage{graphicx}
\usepackage{arydshln}

\usepackage{tabularray}

\usepackage{graphicx}
\usepackage{amsmath,bm}
\usepackage{amsthm}
\usepackage{booktabs}

\usepackage{makecell}
\usepackage{multirow}
\usepackage{ragged2e}
\usepackage{textgreek}
\usepackage{tabularray}

\usepackage{multicol}
\usepackage{algorithm, algpseudocode}

\usepackage{stfloats}

\usepackage[caption=false]{subfig}

\usepackage[normalem]{ulem}
\usepackage{siunitx}

\usepackage[most]{tcolorbox}
\usepackage{listings}

\title{Adversarially Attacking Symbolic Vocabulary Vulnerabilities In LLM Planners}

\author{Stephen Obadinma \& Xiaodan Zhu \\
  Department of Electrical and Computer Engineering, \\ Ingenuity Labs Research Institute, Queen's University \\ \texttt{\{16sco, xiaodan.zhu\}@queensu.ca} }

\begin{document}

\maketitle

\begin{abstract}
As large language models (LLMs) are increasingly deployed as autonomous agents,
their abilities in navigating automated planning tasks have become essential. However, despite their strengths, LLMs' planning ability is not robust, particularly when they are not allowed to exploit commonsense cues in the symbolic vocabulary (e.g., names of actions, predicates, and objects) used to define planning tasks. This makes LLMs unreliable compared to symbolic planners. With recent reasoning models showing improved planning performance on tasks with obfuscated symbolic vocabularies, we utilize an adversarial attack framework to reveal to what extent the crucial weaknesses of LLM agents at joint symbolic interpretation and planning remain and whether they can reason properly when presented with severe cases. As such, our main contribution is devising an LLM-based attack model, which we call \texttt{Symbolic-Swapper}, to automatically generate obfuscated planning domains with non-standard symbolic vocabularies that systematically break strong reasoning models' ability to plan. In doing so, we reveal how much LLMs' planning abilities can be further degraded under more advanced obfuscation schemes, allowing us to ascertain their success when having to rely on their pure planning ability. We find that attacks can decrease the planning performance of base models and even robust planning frameworks by over 70\%. We further analyze the factors behind how LLM planning abilities break down and find that successful obfuscation causes models to appear to significantly degrade in reasoning quality despite them showing some inherent ability to understand the domain, revealing a gap in the perception of a task domain and being able to plan under it.
\end{abstract}

\section{Introduction}

The unprecedented capabilities of LLMs at sequential decision-making and multi-step reasoning \citep{minaee2024largelanguagemodelssurvey, openai2025o4mini} have enabled them to be the basis of autonomous agents in agentic AI frameworks capable of completing planning tasks in complex, open-ended environments \citep{ huang2024understandingplanningllmagents, liu2024agentbench, wei-etal-2025-plangenllms}. These abilities have been critical for their adoption in important planning tasks in embodied robotics \citep{li2024embodied}, web and computer navigation \citep{zhou2024webarena, chae2025web} and automated coding \citep{zhang-etal-2024-codeagent}.
Furthermore, the capacity to plan with just a natural language prompt makes them more versatile and easier to deploy in the real-world compared to rigid symbolic planners. However, despite these advantages, state-of-the-art LLMs exhibit notable limitations in being able to plan robustly that prevent their adoption in high-stakes domains. 
LLM-based agents often struggle with maintaining coherent long-term plans \citep{sun2023adaplanner, gonzalez-pumariega2025robotouille}.
As such, they are frequently subject to compounding errors that they struggle to revise \citep{chen2025scaling, hao2025planning}, and can exhibit myopic or inconsistent strategies, which are difficult to interpret \citep{valmeekam2023on, zhu2024knowagent}, hindering the ability to steer their behaviour in a controlled manner. Moreover, LLMs often have to rely on external scaffolding for optimal planning performance \citep{sel2025llms}.

One of the most fundamental limitations of LLMs is their sensitivity to the symbolic vocabulary defined in a task \citep{valmeekam2023planbench, valmeekam2023on,valmeekam2025a}. Existing LLM planners exploit commonsense cues in task descriptions rather than robustly reason across any possible vocabulary, unlike conventional symbolic planning systems, such as satisfiability modulo theories (SMT) solvers that give deterministic solutions and are invariant to the names used for variables \citep{10.5555/1792734.1792766}. This can make them appear as strong planners on conventional benchmarks despite having limitations in their ability to truly comprehend and manipulate symbolic states. When the symbolic vocabulary is obfuscated, and hence the names that define actions, predicates, and objects become arbitrary, LLMs can struggle despite the task definition remaining fully solvable \citep{valmeekam2023planbench}. 
Most notably, the extent of the fragility of LLMs at joint symbolic understanding and reasoning for planning tasks is unclear. This fragility presents a risk, as any planning tasks not described using semantically meaningful identifiers can become problematic. Additionally, this unreliability in LLMs means that they will not be able to fully replace symbolic solvers in many key tasks, so the myriad advantages of LLMs cannot be leveraged in these cases. There is also the potential for adversaries to derail planning tasks by corrupting symbol names to something unfavourable to a victim model.
Recent planning frameworks attempt to rectify robustness issues by allowing LLMs to use tools and symbolic solvers \citep{hao2025planning}, and recent reasoning models have shown improved performance on benchmarks with obfuscations \citep{valmeekam2025a}, but the underlying vulnerabilities remain.

\begin{figure*}[!t]
  \centering
  \includegraphics[width=0.65\linewidth]{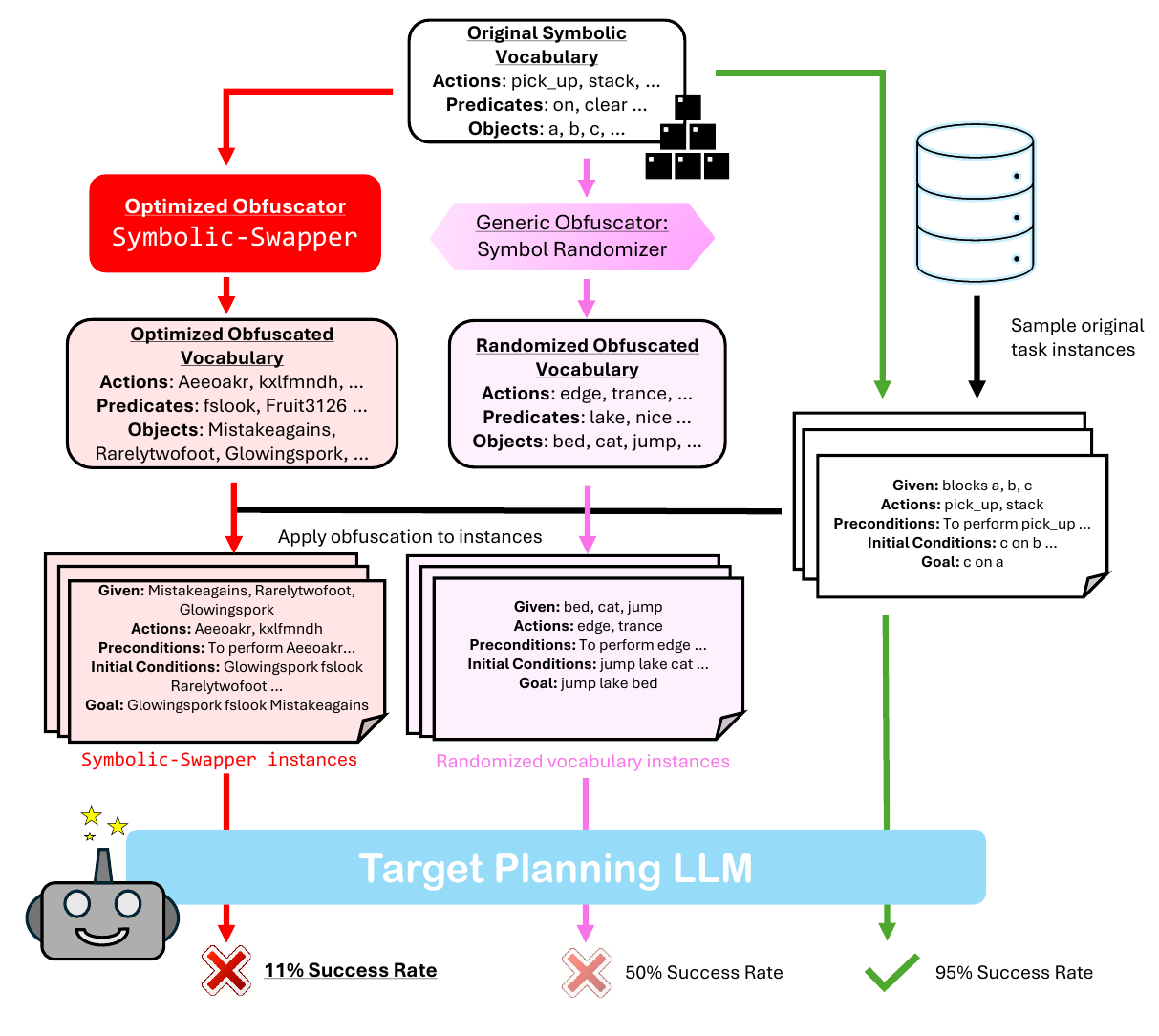}%
  \caption{By using an optimized attack model \texttt{Symbolic-Swapper} to generate a new symbolic vocabulary that is applied to selected task instances, we can reduce plan success rate on a target LLM to extremely low levels compared to their base performance and when using a naive symbol randomizer as an obfuscator. }
  \label{fig:introfigure}
  \vspace*{-1.2\baselineskip}
\end{figure*}

Compared to previous approaches evaluating LLM planners under abstract or obfuscated representations, \textbf{our key objective with this work is to demonstrate how much LLM planning abilities can be further degraded by stress testing them using optimized obfuscation schemes}. We build on prior work by proposing a framework to automatically optimize generating obfuscated representations to successfully break model planning capabilities and analyze the failure modes of LLMs while planning under tricky obfuscated domains. By optimizing adversarial attacks in the context of obfuscated domain planning problems, the core model vulnerabilities in joint symbolic interpretation and planning ability are revealed since models have to rely on their pure planning capabilities, as they cannot exploit patterns in the domain and other background connections to reason around the symbols. This also allows us to discover whether LLMs can robustly reason using consistent symbolic state manipulation. We accomplish this objective by introducing an approach to generate obfuscated names using an attack language model optimized using Group Relative Policy Optimization (GRPO) \citep{shao2024deepseekmathpushinglimitsmathematical} called \texttt{Symbolic-Swapper}. Essentially, we design an adversarial stress test method for automatically generating obfuscated planning problems that cause even strong planners to fail. The overview of our approach is presented in Figure \ref{fig:introfigure}, where we show that using an optimized attack model to create planning instances with an altered symbolic vocabulary induces a high rate of failed plans from a target model. We empirically demonstrate that our attack method has a high success rate and most crucially, generalizes across different architectures. We validate our approach on a variety of tasks under popular planning frameworks. We further motivate our decoupled evaluation of semantics and planning in detail in Appendix \ref{appendix:detailed_motivation}. Our contributions are summarized as follows:

\vspace{-2mm}
\begin{itemize}[noitemsep, leftmargin=20pt]
  \item We \textbf{comprehensively study the vulnerabilities of planning under obfuscated symbolic vocabularies} under a variety of obfuscation styles and planning frameworks.
  \item We show that by using \texttt{Symbolic-Swapper}, a tuned adversarial LLM, the symbolic vocabularies generated by it can \textbf{reduce base planning performance by over 70\%} even on relatively easy planning tasks and on robust planning methods, revealing the extent of the critical lack of robustness in joint symbolic interpretation and planning in LLMs.
  \item By analyzing the factors and trends with the generated plans and reasoning traces, we observe that optimized obfuscated domains induce our tested models to degrade in reasoning quality and self-correction despite innately understanding the domain, demonstrating a gap in how these models perceive a task and their planning ability.
  
\end{itemize}

\vspace{-3mm}
\section{Related Work}
\vspace{-2mm}

\paragraph{Planning Methods:} Improving planning abilities by offering additional scaffolding that LLMs can utilize to reason has been a key objective of recent work. ReACT \citep{yao2023react} established a strong baseline by interleaving reasoning and tool use in a sequential loop, enabling flexible decision-making. Subsequent methods introduce explicit planning structure such as tree-based reasoning frameworks \citep{yao2023tree, choi2025reactree} that cast inference as search over multiple reasoning trajectories with backtracking and evaluation. Other works focus on plan decomposition and refinement \citep{fu-etal-2025-preact} and incorporate reflection and memory mechanisms~\citep{kim-etal-2025-reflact}, emphasizing feedback-driven correction. More recently, formalized planning formulations, such as in \citet{hao2025planning}, frame reasoning as explicit optimization problems to be solved to great effectiveness. We include a variety of these popular planning approaches as baselines and test their vulnerability to symbolic vocabulary attacks.

\vspace{-1mm}
\paragraph{Adversarial Attacks Against Planning Models:}

Recent work exposes the vulnerabilities of LLM-based planning frameworks under adversarial manipulations. \citet{mo2024tremblinghousecardsmapping} explores the overall adversarial threat to LLM agents along different attack strategies. \citet{zhang-etal-2025-breaking} demonstrates that attackers can exploit the inherent instability of multi-step planning by inducing agents to execute repetitive or irrelevant actions. Complementarily, \citet{yang2024watch} shows that agentic systems introduce new backdoor attack surfaces beyond standard LLM settings, including triggers embedded in intermediate observations. 
Likewise, other works study the safety of web-based agents against attacks through multiple mediums such as backdoors and environmental injections \citep{tian2023evil, wang-etal-2024-badagent, wu2025dissecting, liao2025eia}. 
\citet{xu2025advweb} specifically train an adversarial prompter using reinforcement learning to manipulate a target web agent. Attacks on embodied systems test diverse attack scenarios such as optimized jailbreak-based attacks \citep{shuyuan2024exploring, zhang2025badrobot}. We prioritize studying vulnerabilities with symbolic vocabulary by optimizing an adversarial prompt generator. 

\vspace{-1mm}
\paragraph{Planning Under Obfuscated Domains:}
\citet{valmeekam2023planbench} show the significant degradation in planning performance by using non-commonsense symbols in PDDL-based planning tasks. However, they are limited in how they generate obfuscated domains, primarily relying on symbol randomizers. Other work further critically examines the planning capabilities of LLMs under obfuscated domains \citep{valmeekam2023on, Huang_Cohn_Lipovetzky_2025}. \citet{hao2025planning} show that performance can be improved under these domains by formulating planning tasks as optimization problems. Unlike these previous approaches, we craft an explicit optimization process to create obfuscated domains that are designed to test the limits of how much LLMs' planning performance can be degraded. 

\section{Obfuscated Domain Optimization}
\label{section:methods}

The central objective of our study is to investigate how severely LLM planning capability can be degraded by generating adversarially optimized obfuscated domains with an altered symbolic vocabulary. We specifically choose to focus on formal and discrete planning problems, in which an LLM agent must generate a sequence of valid actions that transforms an initial world state into a state satisfying a specified goal. Formally, let $\mathcal{A}$ denote a finite action set from which the agent can perform actions from, $\mathcal{P}$ a set of predicates that define the state of the world by describing properties of objects or relationships between them, and $\mathcal{O}$ a set of objects that can potentially be interacted with and entities that may appear in the environment (such as an agent). Each state $s \in \mathcal{S}$ is represented as a collection of grounded facts over $\mathcal{P}$ and $\mathcal{O}$ and may optionally include object-level descriptors such as colour or other relevant attributes from a finite set $\mathcal{D}$ that allow specific differentiation of objects (e.g., to define a specific goal object).

A planning instance (or query) is specified by an initial state $s_0$, a goal condition $g$, and action semantics that define how actions in $\mathcal{A}$ change the world state. Each action $a \in \mathcal{A}$ is associated with a symbolic schema, typically consisting of parameters, preconditions, and effects. When grounded over objects in $\mathcal{O}$, an action becomes applicable in state $s$ if its preconditions are satisfied, and its execution induces a transition to a successor state $s'$. Furthermore, predicates define the preconditions and effects allowed. Each action $a \in \mathcal{A}$ and predicate $p \in \mathcal{P}$ can act on any number of objects. A valid plan is therefore a sequence of grounded actions $\pi = (a_1, \dots, a_T)$ such that successive application of the actions from $s_0$ yields a terminal state $s_T$ that satisfies $g$. Collectively, $\mathcal{A}, \mathcal{P}, \mathcal{O}$, and, if relevant, $\mathcal{D}$, constitute the primary symbolic vocabulary of the planning task. As the definitions of the schema defining the preconditions and effects are also written in terms of arbitrary symbols, a task is fully solvable irrespective of the names chosen for the symbols; hence, a symbolic solver would still be able to generate and verify a plan.

\subsection{Attacks Against Symbolic Vocabulary}
\label{section:obf_domain}

\begin{figure*}[!t]
  \centering
  \includegraphics[width=0.65\linewidth]{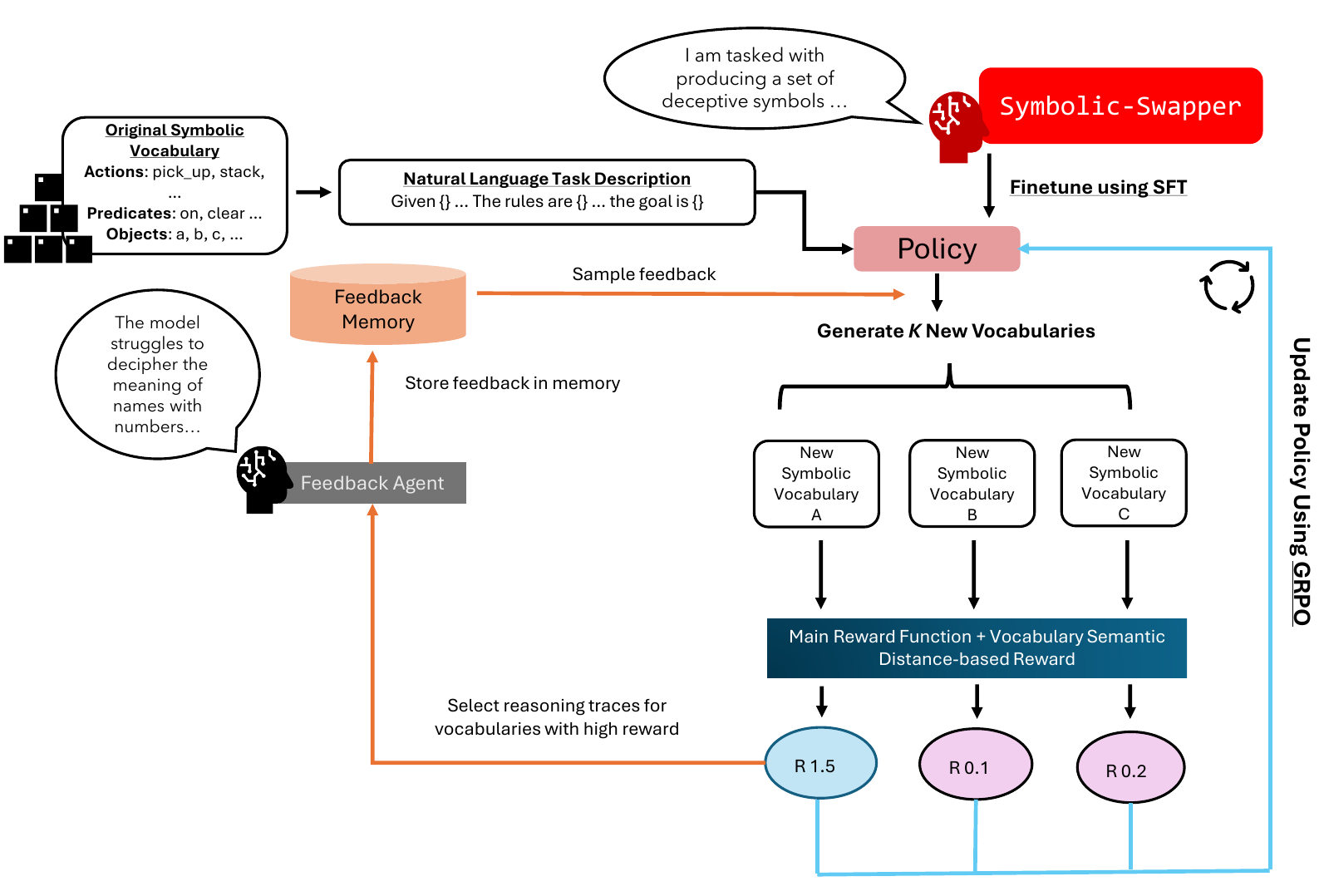}%
  \caption{Optimization process for \texttt{Symbolic-Swapper}. An attack language model is conditioned on a task description and prompted to generate a deceptive symbolic vocabulary. The attack model policy is optimized using GRPO. A semantic distance-based reward added to the main reward function and a feedback agent help to guide policy optimization. }
  \label{fig:introfigure2}
  \vspace*{-1.1\baselineskip}
\end{figure*}

\paragraph{Objective:} Given a task with a defined symbolic vocabulary, we craft an obfuscated version of the domain by mapping all of the symbols from their original meaningful versions to alternate versions that do not necessarily have any inherent meaning or relevance to the original context. The intuition is that under an obfuscated domain with counterintuitive names, LLMs struggle, as common patterns in language cannot be exploited (e.g., such as \textit{picking up} an object typically being followed by \textit{holding} it). The model must infer the latent symbolic structure of the task and produce a sequence of actions that is both syntactically well-formed and semantically consistent. Formally, we define an arbitrary \textit{mapping function} $f$ that maps the symbolic vocabulary (i.e., the set of symbols) \mbox{$\Gamma = \{$ $\mathcal{A}$, $ \mathcal{P}$, $\mathcal{O}$, $\mathcal{D}$ $\}$} to a new form  $f(\Gamma) \rightarrow \hat{\Gamma}$. An obfuscated planning instance $\hat{t}$ is formed by taking an original planning instance for a task and replacing each mention of a symbol with its equivalent in $\hat{\Gamma}$.
The \textbf{objective} therefore becomes to optimize the mapping function $f$ such that the probability of generating a plan $y$ by the target LLM that reaches the goal condition $g$ is minimized for an instance $\hat{t}$ under the new symbolic vocabulary $\hat{\Gamma}$. 
More precisely, given a function $H$ that takes a vocabulary $\hat{\Gamma}$ and task instance $t$ and maps it to a new natural language task instance $\hat{t}$, we aim to optimize $f$ to generate a dataset of obfuscated planning instances:
$
\hat{D} = \{H(f(\Gamma),t_i)\}_{i=1}^{n},
$
 such that we minimize the probability of generating a successful plan \(y\) from a target LLM:
$
\min_{y} \; \mathbb{E}_{\hat{t} \sim \hat{D}}
\left[
P\bigl(\text{success} \mid y, \hat{t}\bigr)
\right]
$
where $\hat{t} = H(f(\Gamma),t)$. The success of the generated plan is obtained according to a symbolic verifier conditioned on the task domain. 

The substitution process can be performed using a multitude of mapping functions (i.e., obfuscators). The simplest is where each symbol in the symbolic vocabulary is mapped to a random English word or a sequence of randomly generated characters. Nonetheless, more sophisticated and deliberately tricky mapping functions are possible that can drastically increase the failure rate of models by exploiting intrinsic pitfalls. Namely, symbols with counterintuitive semantics and those that exploit innate model vulnerabilities to adversarial text can further degrade planning performance, and hence why we choose to focus on creating a mapping function based on an attack LLM optimized to generate strong obfuscations. 
The only restriction on the new mapped symbols is that they must be a single string unbroken by whitespace. We highlight that the mapping function only targets replacing the symbols and leaves the rest of the task description  unperturbed so that only the influence of the symbolic vocabulary is taken into account. We are not strictly concerned about detectability,
since we primarily wish to stress test powerful reasoning models to their limits.

\begin{algorithm}[h]
\small
\caption{GRPO-Based \texttt{Symbolic-Swapper} Optimization}
\label{aglo:main}
\label{alg:grpo}
\begin{algorithmic}[1]
\Require Policy $\pi_\theta$, reward function $r()$, dataset of planning instances $D$, dataset of effective randomized symbolic vocabularies $D_{\mathrm{SFT}}$, I iterations, attack prompt $q$, original symbolic vocabulary $\Gamma$, vocabulary to task mapping function $H()$, semantic similarity model $\text{SIM}()$, negative reward function NegativeReward(), group size $K$, clip parameter $\epsilon$, KL coefficient $\beta$, scaling term $\kappa$
\State Optimize $\pi_\theta$ using supervised fine-tuning on $D_{\mathrm{SFT}}$  \Comment{Minimize the autoregressive negative log-likelihood}
\State $\pi_{\mathrm{ref}} \leftarrow \pi_\theta $ \Comment{Initialize reference policy}
\State Feedback Memory $ \rightarrow []$
\For{I iterations}
    \State $m \gets \text{UniformRandom}( \text{Feedback Memory}) $ \Comment{Draw a simple random sample from Feedback Memory}
    \State $x \gets q + m $ \Comment{Concatenate $q$ and $m$ into one input prompt }
    \State Sample $\{\hat{\Gamma}_i\}_{i=1}^K \sim \pi_\theta(\cdot \mid x)$ \Comment{Sample $K$ responses using policy with $x$ as an input }
    
    \State Sample task instance $t \sim D$
    \For{$i = 1$ to $K$}
        \State $z_i \gets$ $H(\hat{\Gamma}_i,t)$ \Comment{Apply symbolic vocabulary from $\hat{\Gamma}_i$ to $t$}
        \State $r_i \gets r(z_i)$  \Comment{Generate reward based on target LLM plan success}
        \State $r_{sem_i}  = 0$
        \For{symbol $j$ in $\hat{\Gamma}_i$} \Comment{Positive reward for greater semantic distance}
        \State $v \gets $ find corresponding symbol for $j$ from $\Gamma$
        \State $r_{sem_i} += (1 - \text{SIM}(v,j))*\kappa$  \Comment{Positive reward based on semantic similarity function}
        \EndFor
        \State $r_{sem_i}  = r_{sem_i} / \text{total \# symbols in } \hat{\Gamma}_i$ \Comment{Normalize by total number of symbols}
        \State $r_{neg_i} \gets \text{NegativeReward}(\hat{\Gamma}_i)$  \Comment{Generate negative reward based on formatting errors}
        \State $r_i = r_i  + r_{sem_i} + r_{neg_i}$ \Comment{Sum all rewards and penalties}
    \EndFor
    
    \State $\mu \gets \frac{1}{K} \sum_{i=1}^K r_i$
    \State $\sigma \gets \sqrt{\frac{1}{K} \sum_{i=1}^K (r_i - \mu)^2 + \epsilon}$
    
    \For{$i = 1$ to $K$}
        \State $A_i \gets (r_i - \mu)/\sigma$ \Comment{Normalize rewards to obtain advantages}
        \State $\rho_i \gets \frac{\pi_\theta(\hat{\Gamma}_i \mid x)}{\pi_{\mathrm{ref}}(\hat{\Gamma}_i \mid x)}$
        \State $\hat{\rho}_i \gets \mathrm{clip}(\rho_i, 1-\epsilon, 1+\epsilon)$
        \State $\ell_i \gets \min(\rho_i A_i,\; \hat{\rho}_i A_i)$
    \EndFor
    
    \State $\mathcal{L}_{\mathrm{GRPO}} \gets \frac{1}{K} \sum_{i=1}^K \ell_i$
    
    \State $\mathcal{L}_{\mathrm{KL}} \gets -\beta \cdot \frac{1}{K} \sum_{i=1}^K 
    \log \frac{\pi_\theta(\hat{\Gamma}_i \mid x)}{\pi_{\mathrm{ref}}(\hat{\Gamma}_i \mid x)}$
    
    \State $\mathcal{L} \gets \mathcal{L}_{\mathrm{GRPO}} + \mathcal{L}_{\mathrm{KL}}$
    
    \State Update $\theta$ via gradient ascent on $\mathcal{L}$
\EndFor
\end{algorithmic}

\end{algorithm}
\vspace*{-1.1\baselineskip}

\subsection{GRPO-Based Optimization: Symbolic-Swapper}
\label{section:symbolic_swapper}

We introduce a novel mapping function for symbolic vocabularies that acts as an adversarial prompt generator that automatically generates strong obfuscated domains using an attack language model optimized using reinforcement learning (RL) called \texttt{Symbolic-Swapper} (see Figure \ref{fig:introfigure2} for an overview). With this approach, we can directly optimize the mapping function on the main attack objective to create effective obfuscated domains that can be used against any target model.

Given an attack language model, we first create a prompt (see in Appendix \ref{appendix:grpo_settings} for details) that defines the task for the attack model. More precisely, it specifies that given a task description for a planning task, the goal is to generate a set of deceptive names for each entry in the symbolic vocabulary all in one output. We chose our attack model to be a small, lightweight, easily trainable model that has the capacity to understand the task but is more easily tunable for minimal cost.
As the chosen attack model is unlikely to inherently know what are effective deceptive symbols, we further optimize the model using Group Relative Policy Optimization (GRPO) \citep{shao2024deepseekmathpushinglimitsmathematical}. This is an RL–based optimization method designed to improve policy learning by leveraging relative performance within groups of sampled actions.
In GRPO, multiple candidate outputs are generated for the same input, and their rewards are normalized or ranked within the group to compute advantage estimates. This relative comparison mitigates issues such as reward scale sensitivity and high variance in sparse or noisy reward settings.
The stable and sample-efficient updates are beneficial when the reward is costly to compute, such as in our case. 

The overall algorithm for the GRPO-based optimization can be seen in Algorithm \ref{aglo:main}, where we sample multiple $K$ different adversarial symbolic vocabularies each time step and generate a scalar reward. The main reward function generates a score by taking as input a randomly sampled planning task instance and one of the $K$ sampled vocabularies and generates a score based on a target LLM that we want to maximize the success of the obfuscations on. This model is used to generate a plan for the newly generated domain, and the reward is primarily based on the binary value on plan correctness, with the maximum obtained for a plan that does not reach the goal state. Thus, the reward directly captures planning effectiveness under the new domain. In addition to the base reward, we add a negative penalty based on formatting mistakes. 

\textbf{Algorithm Enhancements:} We make a series of alterations to the algorithm to improve the attack efficiency. \textbf{(1)} We conduct supervised fine-tuning (SFT) on a dataset of successful adversarial symbolic vocabularies before starting RL. This dataset is generated by using a mapping function that maps symbols to random words, and the most successful substitutions are training data (i.e., those that achieve a high plan failure rate for an obfuscated domain across $n$ sampled task instances). We use cross entropy loss to train with the targets being the obfuscated symbolic vocabulary given the main attack prompt. SFT allows the attack model to better adhere to formatting guidelines and task requirements before undergoing RL, leading to quicker and more effective optimization. \textbf{(2)} To better steer the deceptive symbolic vocabulary generation process, we incorporate a feedback agent in the main GRPO loop. This agent is based on an auxiliary LLM that, for every successful plan failure when generating the reward, has a random chance to generate a piece of feedback that is added to textual memory. This feedback is generated by tasking the auxiliary model to look at the target model's reasoning trace and formulate actionable feedback based on where the target model is struggling, prioritizing recognizing patterns in names that are giving the target model issues. When the attack model samples a new symbolic vocabulary, a random piece of feedback is selected from textual memory and appended to the attack task prompt to directly highlight areas of failure to prioritize when generating deceptive names. \textbf{(3)} We add a reward based on semantic distance using a word encoder to maximize the distance between original names and their substitutes to promote the attacker mapping to semantically unrelated words. Formally, given the original symbol $c$ and the mapped version $\hat{c}$, we add a positive reward to the main reward based on a semantic similarity function $(1 - \mbox{Sim}(\mathbf{c}, \mathbf{\hat{c}}))*\kappa$, where $\kappa$ is a scaling term, that produces a positive scalar score that is higher for lower similarity. We validate all of these design decisions in Section \ref{section:ablation}. More details on the algorithm and reward structure can be seen in Appendix \ref{appendix:grpo_settings}.


\vspace{-3mm}
\section{Experiment Setup}
\label{section:experiment_setup}

\paragraph{Tasks:} We choose popular planning tasks that are relatively straightforward for strong planning models. In this way, task complexity does not distract from evaluating their vulnerabilities in terms of symbolic vocabulary. The first two tasks we select are from Minigrid \citep{minigrid2023}, which contain a suite of tasks based on navigating a grid-world environment. The first task we choose is \textit{Go To Object} (GTO), which is a simple task aimed at getting an agent to interact with a specified object. The second task is \textit{Unlock}, which requires the agent to first pick up a key that it then has to use to open a locked door. Lastly, we choose a popular task from the PlanBench benchmark \citep{valmeekam2023planbench}, \textit{Blocks World} (BW), which involves assembling vertical stacks of blocks. 

\vspace{-0.5em}
\paragraph{Obfuscation Methods / Mapping Functions:} 
\begin{itemize}[noitemsep, leftmargin=10pt]
\vspace{-0.5em}
  \item \textbf{Random-Swapper (R-Sw)}: Map each symbol in the symbolic vocabulary to a random English word.
  \item \textbf{Base-Swapper (B-Sw)}: Use an unoptimized attack LLM prompted to generate deceptive names yielding new symbolic vocabularies. 
  \item \textbf{Symbolic-Swapper (Sy-Sw)}: A GRPO-optimized attack LLM prompted to generate adversarial symbolic vocabularies.
\end{itemize}

For all obfuscation methods we deploy explicit safeguards to ensure that the target domain remains fully valid and solvable. Any duplicate or invalid names in the domain are explicitly avoided by either re-sampling or by replacement with a generic numbered name depending on the baseline. We also verify each newly generated instance using a symbolic solver prior to being evaluated by the target models. For further details, refer to Appendix \ref{appendix:detailed_experiment_setup}.

\paragraph{Planning Methods:} The following are popular planning methods and frameworks that we test to determine planning effectiveness.
\vspace{-1mm}
\begin{itemize}[noitemsep, leftmargin=10pt]
  \item \textbf{Zero-Shot Prompting (ZS)}: Simply provide the task description / planning instance to the model and generate the plan all in one step without any demonstrations.
  \item \textbf{Backprompting (BP)} \citep{valmeekam2025a}: Using a set number of iterations (in our case three), we first let the LLM produce a plan and verify its correctness. If it fails, we use the feedback from a verifier as additional context in the next iteration to allow the LLM to correct their unsuccessful planning attempts. 
  \item \textbf{ReACT} \citep{yao2023react}: A framework that combines reasoning and acting with LLMs in an interleaved manner. Across a set number of iterations, the LLM is prompted to generate both verbal reasoning traces and the next action. After generating an action, it receives an observation of the environment, after which it continues the thinking and acting process. This creates a synergistic effect that endows the model with the ability to dynamically reason.
  \item \textbf{LLMFP} \citep{hao2025planning}: A strong zero-shot planning framework whereby an LLM is first prompted to encode a planning problem into an optimization problem, without any task-specific examples. It is then tasked with iteratively generating codes to solve the optimization problem which is paired with a formal planner to verify the success of the executed codes. On failure the model is prompted to self-assess its responses. 
\end{itemize}

\paragraph{Models:} To assess attack performance, we primarily test \texttt{Deepseek-R1-0528} \citep{Guo_2025} as the target model, due to its highly effective planning performance achieving the best performance on leaderboards \citep{valmeekam2023planbench}. To test transfer attacks on alternate architectures, we utilize another popular model \texttt{o4-mini} \citep{openai2025o4mini} owing to its high reasoning capabilities. We also include additional results for \texttt{Qwen3-Next-80B-A3B-Thinking} \citep{qwen3technicalreport} and \texttt{gpt-oss-120b} \citep{openai2025gptoss120bgptoss20bmodel} in Appendix \ref{appendix:new_model_results} to show generalization across different architectures. Due to significant costs, we use \texttt{Deepseek-V3.2 Reasoner} \citep{deepseekai2025deepseekv32pushingfrontieropen} to derive rewards for the GRPO as an approximation to R1. Lastly, for the attack LLM used for the B-Sw and Sy-Sw obfuscation generators, we use \texttt{Llama-3.2-3B-Instruct} \citep{llama3modelcard} due to its lightweight nature but high capacity to follow instructions without requiring any thinking mode. LoRA \citep{hu2022lora} is used to fine-tune the attack LLM.

See Appendix \ref{appendix:detailed_experiment_setup} for a more detailed description of methods and tasks.

\section{Results}
\label{section:results}
\vspace{-2mm}

\paragraph{Vulnerability to Optimized Obfuscated Domains: } Our primary research questions are (1) how vulnerable are LLM planners to optimized symbolic vocabularies (generated by Sy-Sw) and (2) to what extent do LLMs fail at joint symbolic understanding and planning? We also wish to determine the effectiveness of our attack compared to the naive baselines (R-Sw and B-Sw) and the original task domains (None). We evaluate planning performance under all the tested obfuscation methods across all the planning tasks, planning methods, and models with the main results in Table \ref{table:main_results}. Our primary evaluation metric is the planning success rate under the obfuscated domain, and we wish to minimize this. For each original task, we generate 250 instances to get baseline planning performance on the regular domain. For each generated obfuscated domain, we randomly generate five instances for each domain such that the total number of instances tested is 250. Our results reveal that the performance on the original task domains is very high, particularly when using planning methods such as ReACT. Nevertheless, performance universally drops after having to plan under obfuscated domains, particularly in the ZS cases. Sy-Sw is by far the most effective obfuscation method, decreasing the planning performance of the ZS baseline and the robust planning frameworks by 70\%. This demonstrates that LLMs still have severe vulnerabilities in joint symbolic understanding and planning and cannot, in most cases, robustly plan even on ``easy'' tasks with these optimized domains.

\begin{table*}[!t]
\centering
\caption{Effectiveness of different obfuscation methods on plan performance across the tested planning frameworks utilizing both Deepseek-R1 and o4-mini. Note the low planning performance in nearly every case after obfuscation, particularly when using (Sy-Sw). }
\label{table:main_results}
\resizebox{0.7\linewidth}{!}{%
\begin{tabular}{lccccc|l|cccc}
\toprule
 &  & \multicolumn{4}{c|}{\texttt{Deepseek-R1}} &  & \multicolumn{4}{c}{\texttt{o4-mini}} \\ 
\midrule
 &  & \textbf{None} & \textbf{R-Sw} & \textbf{B-Sw} & \textbf{Sy-Sw} &  & \textbf{None} & \textbf{R-Sw} & \textbf{B-Sw} & \textbf{Sy-Sw} \\ 
\cmidrule(lr){2-11}
\multirow{4}{*}{\rotatebox[origin=c]{90}{Unlock}} & ZS & 0.83 & 0.73 & 0.63 & 0.14 &  & 0.93 & 0.57 & 0.70 & 0.16 \\
 & BP & 0.85 & 0.75 & 0.66 & 0.27 &  & 1.00 & 0.85 & 0.93 & 0.32 \\
 & ReACT & 0.99 & 0.92 & 0.95 & 0.53 &  & 0.99 & 0.85 & 0.97 & 0.40 \\
 & LLMFP & 0.98 & 0.93 & 0.83 & 0.03 &  & 0.80 & 0.73 & 0.67 & 0.18 \\ 
\hdashline
\multirow{4}{*}{\rotatebox[origin=c]{90}{GTO}} & ZS & 0.97 & 0.94 & 0.57 & 0.29 &  & 1.00 & 0.94 & 0.77 & 0.60 \\
 & BP & 1.00 & 0.97 & 0.90 & 0.74 &  & 1.00 & 1.00 & 0.94 & 0.76 \\
 & ReACT & 1.00 & 0.99 & 0.97 & 0.92 &  & 1.00 & 0.85 & 0.91 & 0.87 \\
 & LLMFP & 0.93 & 0.87 & 0.83 & 0.20 &  & 0.95 & 0.80 & 0.77 & 0.67 \\ 
\hdashline
\multirow{4}{*}{\rotatebox[origin=c]{90}{BW}} & ZS & 0.89 & 0.56 & 0.55 & 0.33 &  & 0.92 & 0.44 & 0.72 & 0.40 \\
 & BP & 0.92 & 0.80 & 0.96 & 0.76 &  & 0.96 & 0.52 & 0.80 & 0.44 \\
 & ReACT & 0.64 & 0.24 & 0.36 & 0.16 &  & 0.84 & 0.47 & 0.49 & 0.20 \\
 & LLMFP & 0.72 & 0.36 & 0.32 & 0.30 &  & 0.20 & 0.19 & 0.16 & 0.12 \\
\bottomrule
\end{tabular}
}
\vspace*{-1.2\baselineskip}
\end{table*}


\textbf{Transfer:}
One key objective we have is to determine whether domains generated by Sy-Sw generalize to other architectures. Despite being trained based on the rewards from Deepseek-V3.2, Table \ref{table:main_results} shows that Sy-Sw attacks transfer to Deepseek-R1 and also to a completely different architecture in o4-mini, although usually to a less severe extent.
In particular, performance in ZS on the Unlock task is halved, showing that the optimized names generalize across different models.  We confirm that attacks transfer to Qwen3-Next-80B-A3B-Thinking and gpt-oss-120b in Appendix \ref{appendix:new_model_results}. Overall, these results indicate that there are common vulnerabilities in how models reason across symbol semantics.

\vspace{-1em}
\section{Analysis}
\label{section:analysis}
\vspace{-2mm}
In this section we study how models fail under optimized obfuscated symbolic vocabularies.

\vspace{-2mm}
\subsection{Plan Failure Mode and Symbol Patterns}
\label{section:plan_failure_modes}
\vspace{-4mm}

\noindent\begin{wraptable}{r}{7.0cm}
\vspace{-2mm}
\centering
\caption{Results for LLM plan failure analysis. Average plan length (PL), \# self-correction tokens (\#SCT), and reasoning failures (RF) can be seen for the ZS Deepseek-R1 planner on GTO and BW. We compare failed attacks (F. Atk.) to ones where there was a high plan failure rate (S. Atk.). We include the results on the original symbolic vocabulary (O) for reference. }
\label{table:plan_length_properties}
\resizebox{\linewidth}{!}{%
\setlength{\tabcolsep}{3pt}
\begin{tabular}{lcccc|c|ccc} 
\toprule
 &  & \multicolumn{3}{c|}{GTO} &  & \multicolumn{3}{c}{BW} \\ 
\cmidrule(lr){3-9}
 &  & \textbf{PL} & \textbf{\# SCT} & \textbf{RF} &  & \textbf{PL} & \textbf{\# SCT} & \textbf{RF} 
\\ 
\midrule
\multirow{3}{*}{\rotatebox[origin=c]{90}{R-Sw}} & F. Atk. & 12853.5 & 37.8 & 0.12 &  & 27817.0 & 64.4 & 0.07 \\
 & S. Atk. & 21172.8 & 2.3 & 0.95 &  & 22521.9 & 55.1 & 0.58 \\
 & Overall & 15757.6 & 27.9 & 0.38 &  & 25699.0 & 60.7 & 0.12 \\ 
\midrule
\multirow{3}{*}{\rotatebox[origin=c]{90}{B-Sw}} & F. Atk. & 12973.9 & 38.4 & 0.05 &  & 21154.8 & 60.9 & 0.10 \\
 & S. Atk. & 19445.4 & 27.2 & 0.53 &  & 20901.2 & 53.2 & 0.43 \\
 & Overall & 15586.8 & 33.8 & 0.28 &  & 21709.0 & 56.0 & 0.24 \\ 
\midrule
\multirow{3}{*}{\rotatebox[origin=c]{90}{Sy-Sw}} & F. Atk. & 12156.0 & 33.6 & 0.00 &  & 34388.7 & 68.2 & 0.00 \\
 & S. Atk. & 11629.4 & 3.9 & 0.80 &  & 30543.4 & 42.9 & 0.36 \\
 & Overall & 10632.3 & 19.1 & 0.33 &  & 30641.1 & 51.9 & 0.29 \\ 
\midrule
\begin{sideways} O. \end{sideways} & Overall & 13888.7 & 61.1 & 0.03 &  & 14157.8 & 44.96 & 0.14 \\
\bottomrule
\end{tabular}
}
\vspace*{-1.1\baselineskip}
\end{wraptable}

We wish to investigate plan failure modes under obfuscated domains by examining patterns in reasoning traces. By analyzing the direct plans generated by the models under altered symbolic vocabularies, we can reveal patterns in planning behaviour when trying to process these symbols and why they are failing. 
To do this, we isolate successful (ones where four or more sampled instances caused a failed plan) and failed (ones where one or fewer plans failed) obfuscated vocabularies for each obfuscation method and calculate different metrics for both types (and overall as well). First, we examine average plan length to determine whether the model needs more tokens when the obfuscated domain is harder. Second, we examine the average number of self-correction tokens (e.g., ``wait'') in the reasoning trace to determine how often models are self-correcting their plans (see Appendix \ref{appendix:sc_tokens} for the tokens we track). Lastly, we use an LLM monitor (\texttt{GPT-4.1-mini}) \citep{openai2024gpt4technicalreport} that is prompted to take the reasoning trace and classify whether the model is failing at its reasoning process, meaning the plan reasoning is incomplete, vague, poorly justified (or even non-existent) with us reporting the average rate of failures (see Appendix \ref{appendix:reasoning_quality_prompts} for the prompt). The results for Deepseek-R1 on the GTO and BW tasks using ZS prompting can be seen in Table \ref{table:plan_length_properties}.

We observe how successful attacks typically make the model output longer plans for GTO except for Sy-Sw, which has much shorter plans on average. Likewise, BW has successfully attacked plans to be shorter on average. In GTO and BW, successful obfuscated domains cause the model to output fewer self-correction tokens. In GTO, obfuscation in general reduces the number of self-correction tokens compared to the no attack baseline. This may reflect reasoning failure but can also signal that obfuscated vocabularies make the model less able to ground errors, become less likely to recognize that correction is needed, or become generally more confident in invalid plans.
The average rate of reasoning failures appears higher in successful attacks as well. Interestingly, Sy-Sw does not have the lowest reasoning quality in its successful attacks, which, combined with the lower overall plan length, shows its mechanism for success is different from the other obfuscation types, where models are led to reason decently on a surface level but still output poor plans. Nevertheless, these trends signal that LLMs may fail to reason and rely less on self-correcting with an optimized symbolic vocabulary. In Appendix \ref{section:world_level_patterns}, our results suggest that issues in reasoning under adversarial symbolic vocabularies are not primarily due to excessively high perplexity or extremely long inputs.
Additionally, in Appendix \ref{appendix:example_reasoning_failure} we provide a qualitative case study of a failure mode that our attacks unveil.

\vspace{-2mm}
\subsection{Translation to Commonsense}
\vspace{-6mm}

\noindent\begin{wrapfigure}{r}{6.0cm}

  \centering
  \includegraphics[width=\linewidth]{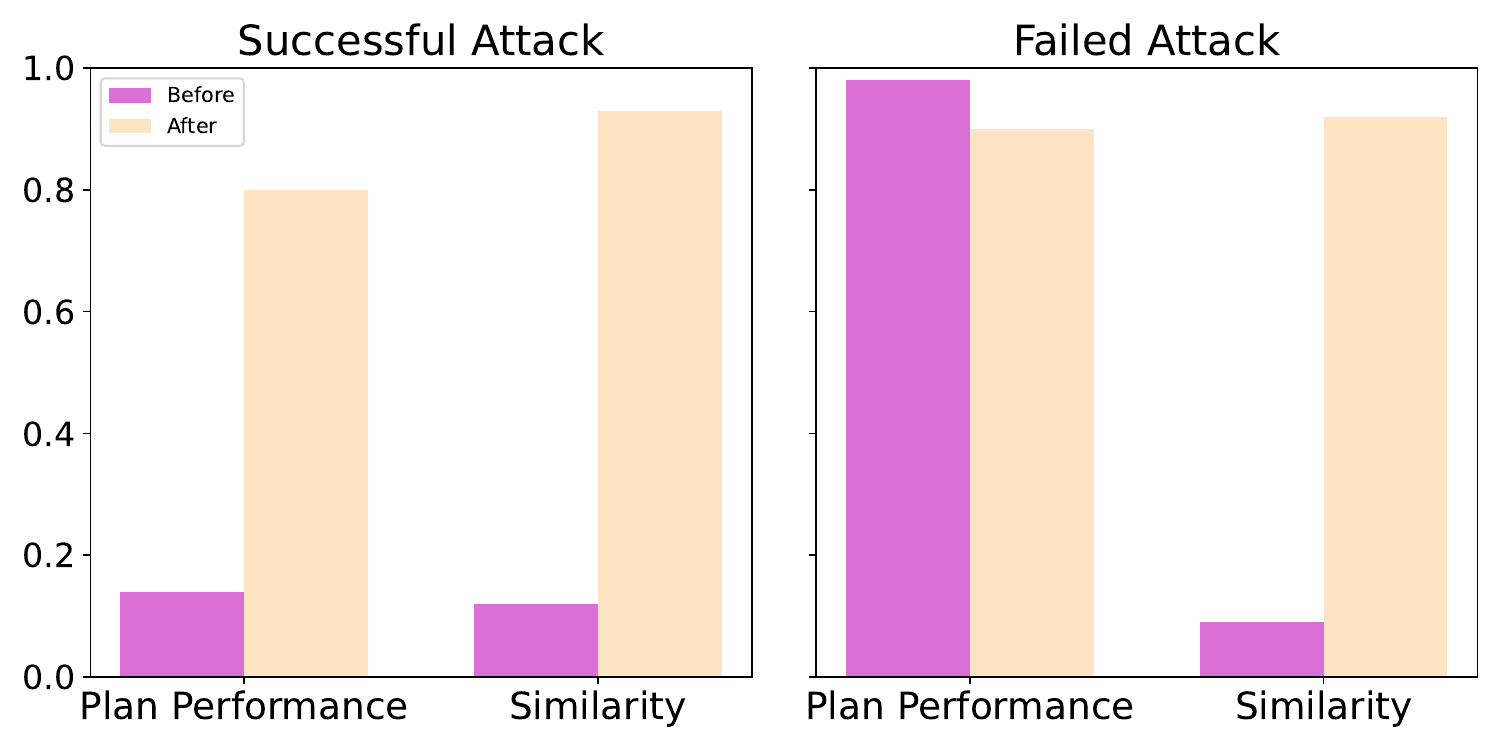}%
  \caption{Result of tasking Deepseek-R1 to translate obfuscated tasks to commonsense on GTO. We report the average plan performance on the translated domain and the average \% similarity between the original symbols and translated versions.}
  \label{fig:figure0}
  \vspace*{-1.5\baselineskip}
\end{wrapfigure}

Considering that models appear to be failing to reason under optimized obfuscated domains, we seek to directly determine whether LLMs can actually determine the intuitive meaning behind the symbols. To do this, we focus on Sy-Sw domains and again separate them into successful and unsuccessful symbolic vocabularies. We take each adversarial domain and task Deepseek-R1 to make its best attempt to recover the commonsense meaning of each symbol based on the context of the task description. We then compare the similarity between the original symbols and the translated ones using GPT-4.1-mini as a monitor LLM that determines whether the function of the name was successfully recovered. Note it does not need to be exactly the same word since we are looking at general meaning and functionality in the context (see Appendix \ref{appendix:similarity_prompts} for the monitor prompts). Using the translated symbols, we use them to construct new task descriptions and collect average planning performance across each. The results can be seen in Figure \ref{fig:figure0}. Despite failing on the obfuscated domains, models figure out the new ``lexicon'' and what is represented by the original names at a high rate of accuracy,  irrespective of whether the symbolic vocabulary deceived them well or not. The LLM then proceeds to generate a high level of plan success after using this translated domain. Essentially, successful obfuscated domains induce models to significantly degrade in reasoning quality despite them innately understanding the domain when prompted to recover it. In this sense, we reveal that there is a gap between how models plan directly based on the names and whether they can actually recover from them. This also signals that LLMs may struggle to plan and reason about the commonsense meaning of the domain at the same time, which causes plan performance to not be robust. These results are generally confirmed for o4-mini and for BW. Although LLMs struggle a bit more in recovering the original domain, they still do it successfully for most of the symbols, as can be seen in Appendix \ref{appendix:translate_results}. 

\vspace{-3mm}
\subsection{Trajectory Analysis}
\label{section:hidden_state}

\begin{figure*}[!t]
  \centering
\subfloat[]{%
  \includegraphics[width=0.32\linewidth,height=3.2cm]{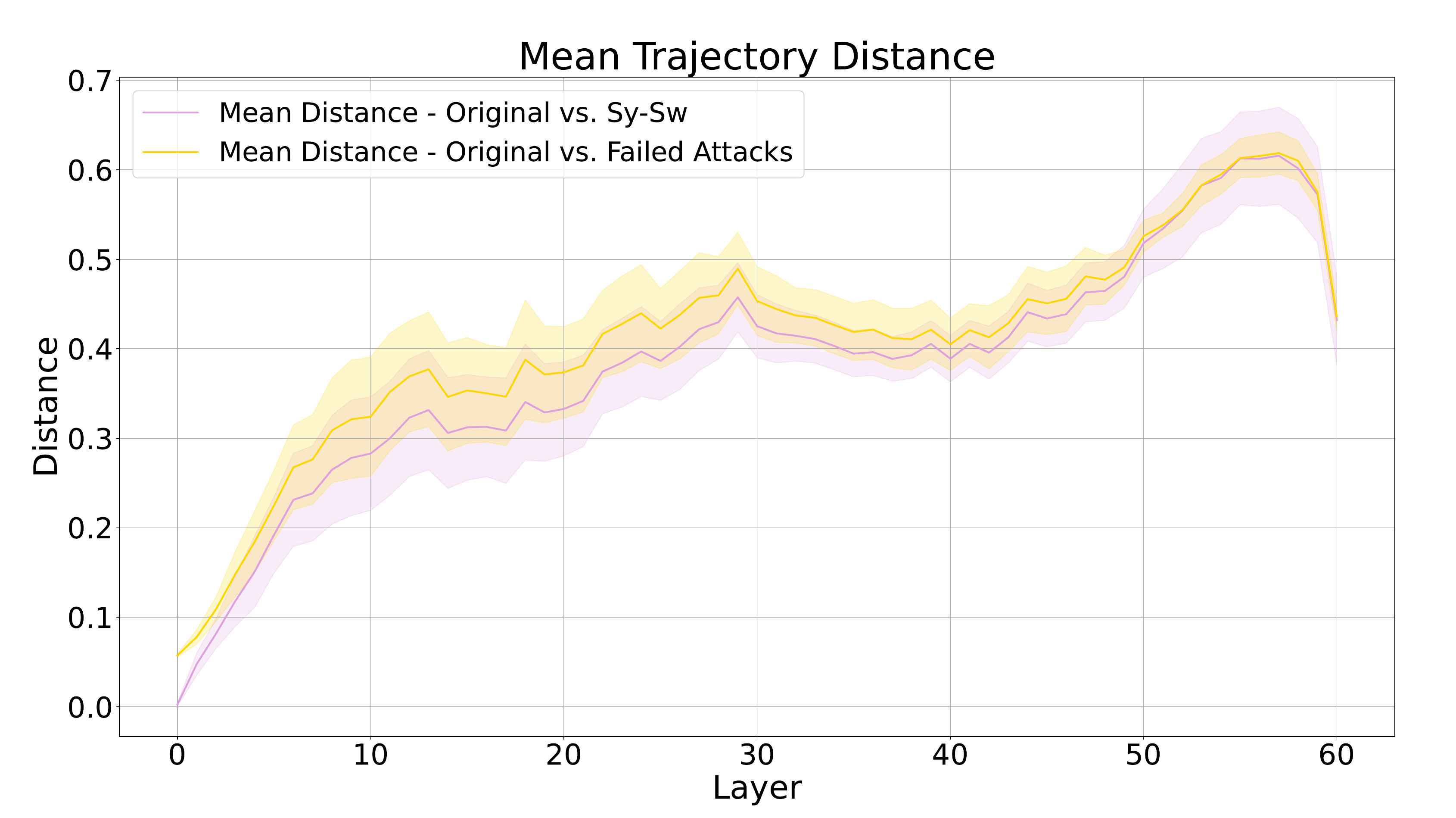}%
  \label{fig:figure1a}
}
 \hfill
\subfloat[]{%
  \includegraphics[width=0.32\linewidth, height=3.2cm]{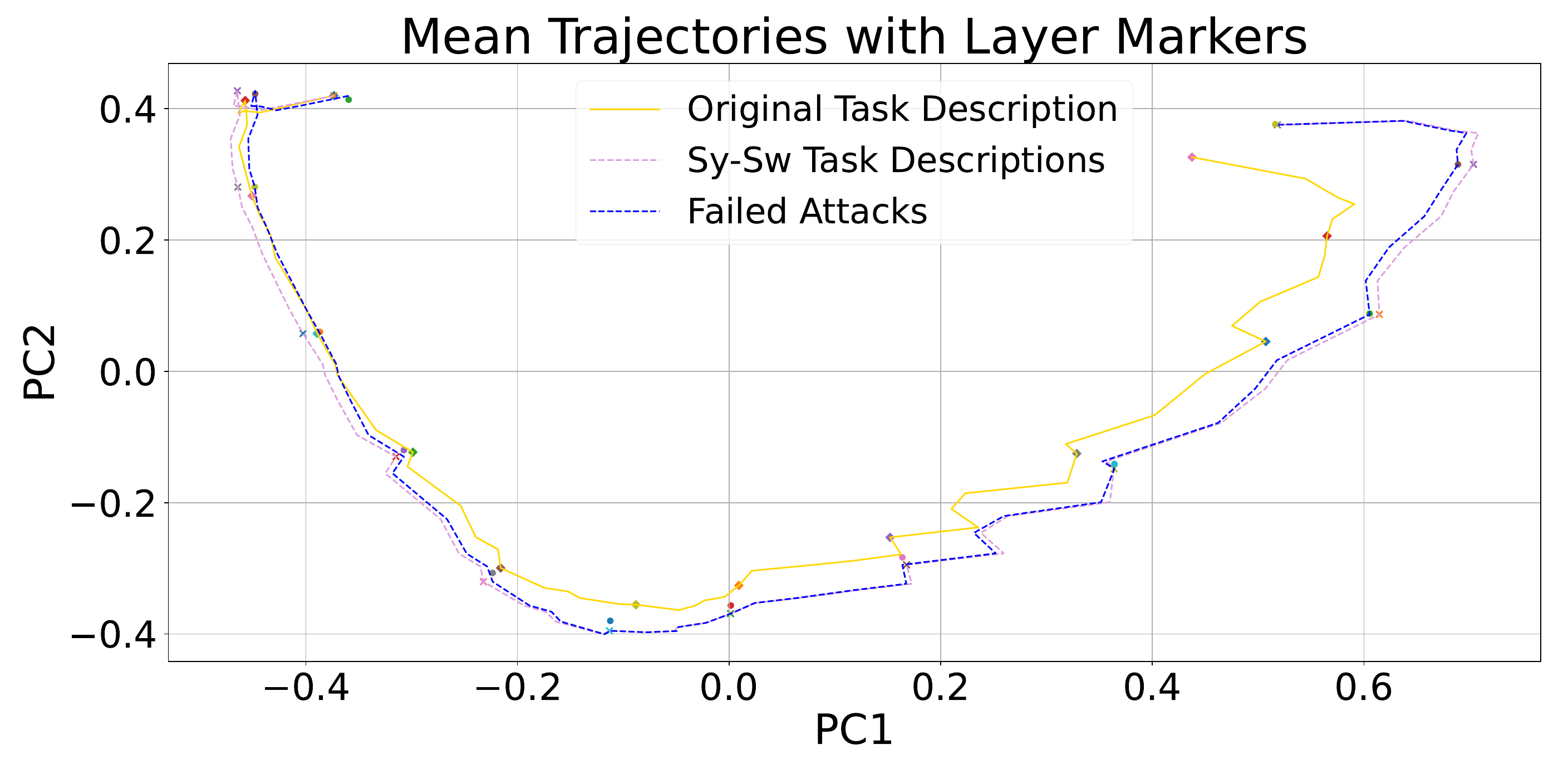}%
  \label{fig:figure1c}
}
\hfill
\subfloat[]{%
  \includegraphics[width=0.32\linewidth, height=3.2cm]{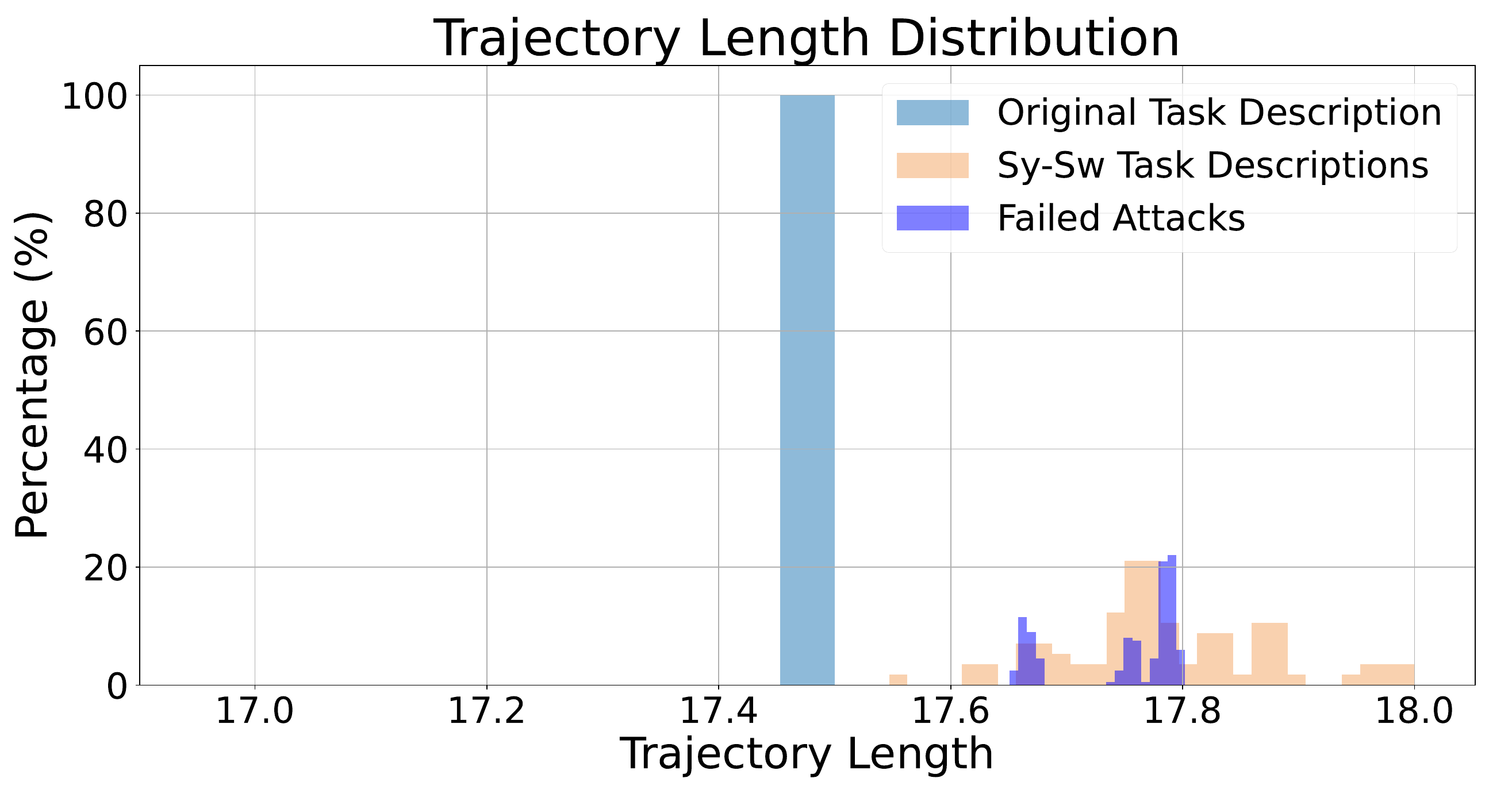}%
  \label{fig:figure1d}
}
\caption{Trajectory analysis on Deepseek-R1 for the GTO task comparing the original task description with the Sy-Sw-generated variants and solely the failed attacks. (a) Mean Trajectory Distance (b)  Mean Trajectories PCA plot (c) Trajectory Length Distribution. }
\label{fig:figure1}
\vspace*{-1.4\baselineskip}
\end{figure*}

To further study how LLMs perceive the optimized obfuscated domains, we conduct a qualitative trajectory analysis based on the hidden states for each layer of Deepseek-R1. Note that this analysis is specific to this model and will not necessarily hold for others, but we believe it can provide some insights into how differently an LLM can perceive obfuscated domains. We provide the original task description and its hidden state vector and compare it with the mean hidden state vector across all of the Sy-Sw task descriptions. As an additional control, we also compare with the subset of task descriptions that did not cause the model to fail at planning (failed attacks). We plot the mean trajectory distance between the original and the other two vectors, the mean trajectories themselves, and the trajectory length distribution. The result for the GTO task can be seen in Figure \ref{fig:figure1}. The hidden state analysis shows significant divergence towards the final layers with most of the middle layers showing relatively stable divergence. The output stage, however, is extremely unstable as mean trajectories peak before dropping rapidly, meaning the planning decisions on the task description are likely forming in these late layers. We also observe how the difference in mean trajectory distance between the original domain and the failed attacks and the original and overall data is greater in early layers. Based on the PCA analysis, we observe the divergence for the trajectories between the original description, the Sy-Sw ones, and the failed attacks, with the obfuscated domains in general showing high curvature towards the end of the path, highlighting the instability in reasoning. The trajectory lengths for the obfuscated domain are in general higher than the original as well, showing that the model is exerting more effort processing the new domain. However, the lengths for the failed attacks are lesser on average, potentially showing the model is reasoning more similarly to the original domain on these instances. Similar results can be found for BW in Appendix \ref{appendix:bw_hidden_tate}. Overall, this signals that LLMs may understand the obfuscated domains similarly, but the reasoning process is unstable when making final decisions.

\vspace{-3mm}
\subsection{Symbolic-Swapper Ablation}
\label{section:ablation}
\vspace{-6mm}

\noindent\begin{wraptable}{r}{7.5cm}
\vspace{-2mm}
\centering
\caption{Ablation results for different variants of the GRPO algorithm: \textit{`None'}, \textit{`SFT\_only'} (SFT-O), \textit{`Feedback\_only'} (FBK-O), \textit{`Distance\_Reward\_only'} (DR-O), and \textit{`Altogether'}. Best results are in bold.}
\label{table:ablations}
\resizebox{\linewidth}{!}{%
\renewcommand{\arraystretch}{1.5}
\begin{tabular}{lccccc} 
\toprule
\multicolumn{1}{l}{} & \textbf{None} & \textbf{SFT-O} & \textbf{DR-O} & \textbf{FBK-O} & \textbf{All} \\ 
\cmidrule(lr){2-6}
ZS & 0.27 $\pm 0.07$ & 0.18 $\pm 0.04$ & 0.20 $\pm 0.03$ & 0.19 $\pm 0.03$ & \textbf{0.14} $\pm 0.02$ \\
BP & 0.46 $\pm 0.12$ & 0.43 $\pm 0.09$ & 0.39 $\pm 0.01$ & 0.31 $\pm 0.03$ & \textbf{0.27 $\pm 0.04$} \\
ReACT & 0.85~$\pm 0.03$ & 0.78 $\pm 0.08$ & 0.70 $\pm 0.07$ & 0.75 $\pm 0.03$ & \textbf{0.53 $\pm 0.09$} \\
LLMFP & 0.34 $\pm 0.12$ & 0.09 $\pm 0.03$ & 0.16 $\pm 0.05$ & 0.15 $\pm 0.05$ & \textbf{0.03 $\pm 0.02$} \\
\bottomrule
\end{tabular}
}
\vspace*{-1.2\baselineskip}
\end{wraptable}

We perform an ablation to show the effects of each algorithmic design choice on Sy-Sw. Focusing on the Unlock task on Deepseek-R1, we train five different variants of the GRPO algorithm: \textit{`None'} (basic GRPO), \textit{`SFT\_only'}, \textit{`Feedback\_only'}, \textit{`Distance\_Reward\_only'}, and \textit{`Altogether'}. Our evaluation is averaged across three training seeds for each variant. We present the average plan success rate across each variant in Table \ref{table:ablations}. The variant with all of the design choices produces the lowest plan success rate overall across all planning frameworks, confirming the effectiveness of the algorithm. 

\vspace{-3mm}
\section{Conclusion and Limitations}
\label{section:conclusion}
\vspace{-2mm}

When faced with an altered symbolic vocabulary optimized using an adversarial objective, we have shown that strong LLM planners have their performance substantially degraded. This reveals that their inherent ability to plan over arbitrary symbols is fragile. This is true even when using specialized planning frameworks that focus their reasoning abilities. Through additional analysis, we show patterns on how models perceive obfuscated domains and show that they fail despite having some intuition of the original meaning of symbols, showing a gap between understanding and reasoning ability that can be exploited. The fact that LLMs have an over-reliance on commonsense naming conventions makes their reasoning inherently brittle, and we encourage future work to focus on building planning approaches that promote inherent invariance that does not break down in reasoning when faced with the obfuscated symbolic vocabularies generated by \texttt{Symbolic-Swapper}. 

Given the query-intensive nature of optimizing attacks and testing plans, which require many tokens at a high cost, we are unable to test a large combination of models and planning tasks. We are limited to only large, strong-performing models that already show a high aptitude in planning under obfuscated domains. In our prior testing evaluating smaller-scale, open-source models often results in failure to plan under any sort of obfuscation; hence, it does not make sense to stress test their already poor performance. Nevertheless, we do show generalizable results across multiple strong model types and three tasks. 
In addition, we focus on tasks presented in natural language, given this is the dominant use case for LLM planners as opposed to strictly formalized language like PDDL. 

{
\small

\bibliographystyle{plainnat}
\bibliography{custom}

}

\newpage


\appendix

\section*{Ethics Statement}

This work proposes a method to automatically generate obfuscated planning domains that can severely diminish the planning performance of LLMs compared to their base performance. The  attack model is untargeted and is transferable between different architectures. It is completely black-box relative to a target model, requiring no details to build a successful attack. However, full white-box access is required to train the adversarial attack model. Once the attack model is optimized, the inference cost from an adversary is minimal to create effective obfuscated domains. As such, our methodology can be used by adversarial actors with access to a planning system to build and present planning environments to the system that can compromise its successful operation. The threat model boundary therefore is in the pre-deployment stage, requiring access to avenues such as APIs or file upload forms to modify existing planning environments. In our research study we use our methodology only to analyze model behaviour and discover vulnerabilities, but we acknowledge that there are dual-use implications to our framework.   

We validate our attack methodology in the context of symbolic planning benchmarks in which everything is fully specified and disconnected from real-world embodied systems. We stress that we do not provide any information to specifically bypass safety guardrails for embodied agents, and our attack framework in its current form is not designed for manipulating physical environments or actuators. Our proposed optimization operates entirely over synthetic symbolic encodings and does not target alignment, safety filters, or instruction-following safeguards.

\section{Importance of Stress Testing Joint Symbolic Understanding and Reasoning}
\label{appendix:detailed_motivation}

Our overall goal is to introduce a controlled diagnostic that isolates one factor—dependence on semantic priors encoded in predicate, object, descriptor, and action names. We emphasize that we do not believe that planning should be evaluated completely independently of the semantics of language, but instead we believe that our methodology allows for an important evaluation of LLM capabilities that cannot be achieved solely through testing performance on conventional planning benchmarks. 

Decoupled evaluation such as ours is a diagnostic methodology and is akin to problems in computer vision where previous work has isolated robustness to blur, occlusion, or adversarial perturbations without claiming perturbations represent natural images \citep{REN2020346}. 
Standard planning benchmarks typically use semantically meaningful identifiers (e.g., pickup, stack, clear), which may allow models to exploit or overly rely on lexical knowledge acquired during pretraining, in addition to performing symbolic reasoning. By replacing these identifiers with semantically neutral but logically equivalent representations, the underlying planning problem remains unchanged while linguistic cues are removed. The resulting performance therefore measures the robustness of reasoning under changes in linguistic realization rather than changes in the planning task itself.

In such a case, an LLM may partially solve the task by exploiting these priors rather than by consistently manipulating the underlying symbolic state. For example, assume there is a comparison between two planners that are equally accurate on a benchmark, but one is able to truly track symbolic states and perform logical search while the other purely relies on lexical associations. In this evaluation both models would appear equivalent, while the former planner is clearly more robust and such robustness would be revealed precisely through the decoupled evaluation provided by our methodology. This is why we believe this decoupling reveals failure modes that are difficult to observe in standard evaluations. The two models may achieve similar accuracy on conventional benchmarks, yet differ substantially in their reliance on semantic priors versus consistent symbolic state manipulation. Conventional evaluations cannot distinguish these behaviours because both succeed when informative predicate names are available. 

Evaluating in a normal semantically grounded setting also presents some limitations. It is possible to keep improving natural language understanding, which will concurrently improve planning performance, but few guarantees can be provided as to whether a model really understands the function of the symbols at a core level. Similarly, it would be difficult to ascertain when LLMs are capable of fully replacing symbolic planners, in the sense that they possess the exact same invariance to symbol semantics. Therefore, if an LLM is shown to be able to plan well truly irrespective of how hard to interpret or intentionally confusing a symbol is, then this demonstrates a high level of model robustness that cannot be claimed merely based on evaluating it in normal semantically grounded settings. We therefore view our benchmark as complementary to standard semantically grounded evaluations rather than a replacement. Together, the two settings provide a more complete characterization of LLM planners by separating overall task performance from robustness of the underlying reasoning process. More broadly, controlled interventions are a standard methodology for diagnosing complex AI systems, such as in the case of verbalized confidence \citep{obadinma2025on}, because they enable attribution of failures to specific capabilities, which is difficult to achieve from end-to-end evaluation alone.

Furthermore, although semantic knowledge is integral for many planning tasks that language models can face, it does not cover all of them. There are many planning tasks that have representation invariance under symbol renaming where classical planners like Fast Downward~\citep{1622559.1622565} are used. In such tasks, we ideally would want to be able to use LLMs instead due to their adaptability. There are many cases where the symbolic vocabulary is generated dynamically, learned automatically, or supplied at runtime, such as during API orchestration, robotics, formal verification, theorem proving, and machine-generated workflows. In these settings, the ability to plan correctly regardless of symbol names is tantamount, and as language models prove more capable at agentic tasks, we should expect them to perform as robustly as symbolic planners without external aid, particularly when they have built-in means of self-correction. Likewise, advanced planning agents can use tools, interact with symbolic planners, use formal APIs, read generated PDDL, and solve unseen domains. These situations frequently contain arbitrary identifiers and hence could be vulnerable to attacks by any adversaries that have the ability to manipulate these identifiers. 

A non-exhaustive list is the following of such tasks is:

\begin{itemize}[itemsep=0pt]
\item Code generation over unknown APIs where an LLM may need to plan using functions it has never seen before with arbitrary names (e.g., that can be automatically generated or suggested whereby an adversary can introduce corruption).
\item  In robot planning from learned operators where symbolic identifiers  for actions and objects are generated automatically from perception or learned abstractions.
\item  When abstract predicates are learned that correspond to latent concepts discovered automatically, a downstream planner should manipulate these symbols without assigning them natural-language interpretations.
\item Suppose an enterprise exposes hundreds of APIs. The LLM receives only machine-readable descriptions with arbitrary names and IDs. Planning consists of chaining actions based solely on type compatibility and preconditions.
\item  Planners for formal theorem provers and scientific experiment planning can define completely arbitrary names. In such problems the onus is on the planner to rely purely on its logical reasoning abilities rather than get bogged down by semantics. 
\item  In enterprise workflow automation tasks like IT ticket routing can use autogenerated UUIDs or workflow definitions.
\end{itemize}

As for the specific threat model, an adversary with API access or some ability to manipulate the arbitrary symbols can do so to compromise the operation of a system. For example, someone using a compromised model trained to recommend variable and function names for coding tasks (e.g., for a coding assistant) can cause target planners to generate subpar code. Alternative possible avenues for adversaries include, prompt injection or configuration attacks whereby an attacker can modify API specifications, workflow definitions, or planning-domain files to systematically rename actions and objects without altering their behaviour, targeting an LLM's dependence on lexical cues. Additionally, compromised middleware or intermediary services could have rewritten object IDs, action names, or tool identifiers by an adversary before they reach the LLM while preserving the execution semantics expected by downstream systems so it is not immediately suspicious.

\section{Detailed Experiment Setup and Implementation}
\label{appendix:detailed_experiment_setup}

In this section we provide additional details for our experiment setup.

\subsection{Obfuscation Baselines}

\paragraph{Random-Swapper:} We iterate through each symbol in the task-specific symbolic vocabulary and map them to a random word from a repository of English words. Our source of English vocabulary is from the NLTK words library \citep{https://doi.org/10.48550/arxiv.cs/0205028}. We also ensure that duplicate names are not used.

\paragraph{Base-Swapper:} We take the base instruct-tuned attack LLM (\texttt{Llama-3.2-3B-Instruct}) and iterate through each symbol in the task-specific symbolic vocabulary. For each symbol, we directly prompt the model to generate a deceptive, confusing, or tricky name consisting of a single string uninterrupted by whitespace that is grounded in the context of the original task description and the previous new symbols that it has generated so far to ensure that duplicates are not produced. We use a sampling temperature of 0.7 and a top\_p of 1.0.

\paragraph{Symbolic-Swapper:} See Appendix \ref{appendix:grpo_settings}.

After generating the new symbolic vocabulary, we substitute the original symbols in the task description with the new counterparts. 

\subsection{Planning Tasks Overview}

\paragraph{Go To Object: }

This is a simple task that takes place in an environment consisting of a small grid world made up of navigable cells with two cells containing different objects with specific descriptors, with one of them being designated as the goal object that the agent must interact with. The objective of the planner is to direct the agent to be near the goal object and perform the `done' action. This task does not require long-horizon thinking from the planner and instead presents itself as a task that any reasonable planning LLM can solve consistently; hence, by attacking the symbolic vocabulary, we can determine how much performance can drop and model reasoning can fail even on an easy task such as this. The main interaction space grid is divided into a 4 x 4 grid of cells. The symbolic vocabulary consists of actions, predicates, objects, and descriptors for objects since multiple instances of the same object can be in the environment but are differentiated by a specified descriptor (i.e., colour) which allows the agent to find the goal. The reward is sparse and is only given when the goal is reached. We make the environment fully observable.

The original action vocabulary is defined as follows: 

\begin{itemize}
    \item \textit{turn\_left} - turn the agent 90 degrees counterclockwise.
    \item \textit{turn\_right} - turn the agent 90 degrees clockwise.
    \item \textit{move\_forward} - move the agent forward one cell.
    \item \textit{done} - if the agent is near the goal object, performing this action finishes the task. 
\end{itemize}

The original predicate vocabulary is defined as follows: 

\begin{itemize}
\item \textit{at} - signifies that the agent is at a specific cell.
\item \textit{face} - signifies that the agent is facing a certain direction.
\item \textit{adjacent} - signifies that a cell is adjacent to another cell.
\item \textit{clear} - signifies that a cell is empty and the agent can move to it.
\item \textit{atObj} - signifies that a cell contains a specific object and the agent cannot move to it.
\item \textit{near} - signifies that the agent is near a cell with an object but is not necessarily facing it.
\item \textit{goalObj} - signifies which object is the goal object.
\item \textit{prevDirection} - specifies that a direction symbol precedes another in the cycle.
\item \textit{nextDirection} - specifies that a direction symbol follows another in the cycle.
\end{itemize}

The original object vocabulary consists of the agent and two objects, which can be a `key,' a `ball,' or a `box,' and both objects can be of the same type in a single instance. Each object is located at specific cells in the grid determined by the specific instance of the problem. To differentiate each object, the descriptor vocabulary consists of six symbols as follows: `blue', `green', `grey', `purple', `red', `yellow'. Each object has only one descriptor, and if a instance of the problem has two of the same object, then each will have a different descriptor. 

The overall task description and a task instance can be seen in the prompt in Figure \ref{fig:gto_none}.

\paragraph{Unlock:}

Unlock is a task that takes place in an environment consisting of a small grid world made up of rooms, walls, a door, and a key, where the goal is for the planner to direct the agent to the goal state. The core challenge is that the agent cannot directly reach the goal, and it must first interact with objects in the correct order, making it a more difficult task than Go To Object. The environment is a grid that is  divided into two  4 x 4 grids of cells separated by a locked door. To complete the goal, the agent must unlock the door, but it first must (1) find and pick up the key (2) navigate to the door, and (3) open the door. The symbolic vocabulary consists of actions, predicates, and objects. The reward is sparse and is only given when the goal is reached. We make the environment fully observable. 

The original action vocabulary is defined as follows: 

\begin{itemize}
    \item \textit{turn\_left} - turn the agent 90 degrees counterclockwise.
    \item \textit{turn\_right} - turn the agent 90 degrees clockwise.
    \item \textit{move\_forward} - move the agent forward one cell.
    \item \textit{pick\_up} - pick up the key (only valid when in an adjacent cell to the key and the agent is facing it).
    \item \textit{toggle} - when the agent holds the key performing this action while in an adjacent cell to the door will complete the task.
\end{itemize}

The original predicate vocabulary is defined as follows: 

\begin{itemize}
\item \textit{at} - signifies that the agent is at a specific cell.
\item \textit{face} - signifies that the agent is facing a certain direction.
\item \textit{adjacent} - signifies that a cell is adjacent to another cell.
\item \textit{clear} - signifies that a cell is empty and the agent can move to it.
\item \textit{atObj} - signifies that a cell contains a specific object and the agent cannot move to it.
\item \textit{carrying} - signifies that the agent is carrying an object (e.g., the key).
\item \textit{front} - signifies that an object is directly in front of the agent.
\item \textit{open} - signifies that an object is in the open state.
\item \textit{closed} - signifies that an object is in the closed state.
\item \textit{locked} - signifies that an object is in the locked state.
\item \textit{goalObj} - signifies which object is the goal object.
\item \textit{prevDirection} - specifies that a direction symbol precedes another in the cycle.
\item \textit{nextDirection} - specifies that a direction symbol follows another in the cycle.
\end{itemize}

The original object vocabulary consists of the agent and two objects: a `door' and a `key' located at specific cells in the grid. The cells and direction symbols do not get obfuscated but are written in generic syntax.

The overall task description and a task instance can be seen in the prompt in Figure \ref{fig:unlock_none}.

\paragraph{Blocks World:} is a classical planning problem defined in PDDL language. In its original form, it gives the task of stacking a set of finite objects (blocks) in a specific configuration with constraints that only one block can be held at a time and only clear blocks can be moved. The symbolic vocabulary consists of actions, predicates, and objects.

The original action vocabulary is defined as follows: 

\begin{itemize}
    \item \textit{pick\_up(x)} - pick up an object
    \item \textit{put\_down(x)} - put a held block onto the table
    \item \textit{stack(x, y)} - place block x on block y
    \item \textit{unstack(x, y)} - remove x from on top of y
\end{itemize}

The original predicate vocabulary is defined as follows: 

\begin{itemize}
\item \textit{on(x, y)} - block x is on block y
\item \textit{ontable(x)} - block x is on the table
\item \textit{clear(x)} -  nothing is on top of x
\item \textit{holding(x)} - the robot is holding x
\item \textit{handempty} - robot hand is empty
\end{itemize}

The original object vocabulary consists of objects named `a', `b', `c', `d', `e' but each block does not necessarily have to be present or used in a given planning problem instance.

The overall task description and a task instance can be seen in prompt in Figure \ref{fig:blocksworld_none}.
The initial state specifies the initial configuration of the environment, and the goal specifies which blocks must be stacked on top of which. A symbolic PDDL verifier is used to confirm the validity of generated plans and whether they accomplish the goal.

\subsection{Planning Frameworks Implementation}

\subsubsection{Zero-Shot} 

For ZS we apply the task prompt with the general description of the task and the special instance details and generate the plan all in one step. We extract the final sequence of actions from the plan using \texttt{GPT-4.1-mini} and evaluate it using the relevant plan verifier for the task domain. 

\subsubsection{Backprompting}

Backprompting closely follows ZS, but instead of a single iteration of generating and validating the LLM plan, upon failure we prompt the model for a new plan but provide its previous failed plan as context (including explicitly stating that this previous plan failed) so that the model can reason about why it failed and build a better plan. This self-correction loop occurs over three iterations if a successful plan is not generated before the final step plan is taken as the LLM's best solution.

\subsubsection{ReACT}

For our ReACT implementation we closely follow the original implementation in \citet{yao2023react}. Given the initial prompt with the initial state and goal, we prompt the LLM to generate a single thought and action pair for the next step of the plan. After generating the action, we extract it and evaluate using a plan validator given the plan steps generated so far. If the action is successfully completed, we finish the task. Otherwise, if the goal state has not been reached, we return an observation of the environment and add it to the prompt as context for generating the next step. In the Minigrid tasks, this observation consists of the agent's current position in the grid since that is the primary dynamic variable (objects do not move). For Blocks World we give all of the predicates that are true at the current time-step. If the previous action was not valid, then we explicitly tell the LLM of this and inform it that the environment has not been changed as a result. This loop continues until the maximum number of time steps is reached, which we set to 20 for all tasks.

\subsubsection{LLMFP}

The LLMFP (LLM-based Formalized Programming) method is a planning framework that uses LLMs not to directly generate plans but to translate planning problems into formal optimization programs and solve them rigorously. It requires no task-specific examples.

LLMFP follows a 5-stage pipeline where inputs from each step are fed to the next step:

\textbf{1. Input Definition}: The following are collected and are used to define the planning problem context. They are used as inputs to the LLM and include:
\begin{itemize}
\item A task description in natural language
\item Background info on tools and APIs. This includes access to symbolic solvers like Z3. 
\item The user query

\end{itemize}

\textbf{2. Definer}: focuses on problem structuring. Given a specific prompt the LLM extracts:
\begin{itemize}
\item Goal(s)
\item  Decision variables
\item Constraints (rules, limits, dependencies)
\end{itemize}

\textbf{3. Formulator (Formal Representation)} - Converts the structured problem into a formal model that is often JSON or symbolic representation. The prompt specifically asks the LLM to specify:
\begin{itemize}
\item Variable types (binary, integer, continuous)
\item Relationships
\item Objective function
\end{itemize}

\textbf{4. Code Generator and Solver Execution} - The LLM is tasked with:
\begin{itemize}
\item Generating code (e.g., for an SMT solver)
\item Executes it using a solver
\item The solver finds an optimal or feasible plan satisfying constraints
\end{itemize}

\textbf{5.  Result Formatter and Self-Assessment} - The LLM is prompted to format a solution into a human-readable plan and perform self-checking / refinement to fix any detected errors. Here the correctness is validated, and the model is able to adjust the formulation if needed.

Steps 4 and 5 are repeated over multiple iterations. For our implementation we run 3 runs of step 4 before providing the final input to step 5. If the process does not succeed with successfully executed code, we return to step 4 again and continue for 5 total iterations of the self-reflection loop. We follow the original implementation of LLMFP closely using their prompts. For Minigrid, due to lack of an SMT solver, we provide the API to test the correctness of a plan.

\section{Symbolic-Swapper Details}
\label{appendix:grpo_settings}

In this section we provide specific details for the optimization process of \texttt{Symbolic-Swapper}, including the reward structure, training settings, and prompts used. 

\paragraph{Semantic Distance Reward:} To calculate the semantic distance reward, we use \texttt{text-embedding-3-small} \citep{OpenAI2024} for embedding the original symbol and the one generated by the obfuscator and normalize the score from -1 to 1 to 0 to 1. We then apply a positive reward addition based on 1 minus the normalized score, promoting a bias towards new names that are more distant from their original counterparts. This score is multiplied by a scaling term value of 0.5 to prevent the semantic distance reward from dominating the main reward around planning success.

\paragraph{Processing Inputs and Negative Reward:} To generate the new symbolic vocabulary, we prompt the attack LLM (Appendix \ref{appendix:attack_prompt}) to provide a Python dictionary with keys being the original names and the values being the substituted deceptive names. After the dictionary has been extracted, it must be processed for valid formatting. For every duplicate, unchanged, or missing symbol in the extracted dict, we apply a negative reward of 0.1. We also replace each symbol like this with a generic name (new\_name plus a number based on the order it occupies in the vocabulary) to avoid having any duplicates generating issues when prompting the target LLM for a plan when calculating the reward. The new names are then swapped directly with their counterparts in the relevant task description. 

\paragraph{Overall Reward Structure:} Given a task instance with a generated obfuscated symbolic vocabulary fed into our target LLM and prompted to complete in a zero-shot fashion, if the plan generated is valid and correct, then the model receives a reward of 0, with the negative reward penalty applied as well if relevant. If the target model fails, we apply a reward of 2.0 onto which the semantic distance reward and the negative reward (if relevant) are added. These are the base rewards, as the GRPO algorithm normalizes the rewards for more stable optimization.

\subsection{GRPO Training Settings}
\FloatBarrier

\paragraph{LoRA parameters:} We use the following parameters for initializing the LoRA weights: r=8, lora\_alpha=16, lora\_dropout=0.05, target\_modules=["q\_proj", "k\_proj", "v\_proj", "o\_proj"], and bias="none".

\paragraph{SFT Training:} We initially optimize the attack model policy $\pi_\theta$ using supervised fine-tuning on a dataset of successful obfuscated symbolic vocabularies $D_{\mathrm{SFT}}$. This dataset was generated by sampling 500 symbolic vocabularies generated by the R-Sw and B-Sw attacks that produced a high percentage (60\% or above) plan failure rate. We train LoRA for 20 epochs on the dataset of adversarial symbolic vocabularies with the above parameters. The training objective is to minimize the autoregressive negative log-likelihood of the tokens. The target model output is the list of obfuscated symbols corresponding to the order specified in the attack prompt. The learning rate is 2e-4 using the AdamW optimizer \citep{loshchilov2018decoupled}.

\paragraph{GRPO values:} We use a value of K=4, batch size 32, $\beta$ = 0.1, a learning rate of 2e-4 using the AdamW optimizer, and a max sampling tokens value of 250. We sample new symbols using generation settings temperature=0.7 and top\_p=1.0. We compute rank-based advantages similar to \citet{choi2026gopopolicyoptimizationusing}, normalized to [-0.5, 0.5] with an $\epsilon=$1e-8 and global\_mix=0.2. We train until convergence.

For compute resources, we use NVIDIA L40S 48GB GPUs to both train and run Sy-Sw and run inference on B-Sw. We use NVIDIA H100 SXM 80GB GPUs to conduct trajectory analysis on Deepseek-R1. For o4-mini we use the OpenAI API \citep{openai2025o4mini}. For running inference with DeepSeek-R1, we make use of APIs through OpenRouter \citep{openrouter2024}. For Deepseek-v3.2, we use the official Deepseek API for queries \citep{deepseekai2025deepseekv32pushingfrontieropen}. 

\subsection{Attack Prompt Example}
\label{appendix:attack_prompt}
\FloatBarrier

The prompt used for our attack LLM is unlike the one for B-Sw. Rather than iterating and generating new symbols one by one, we instead prompt the new symbols all in one step in the form of a Python dictionary, which we then extract using \texttt{GPT-4.1-mini}. 

\paragraph{Inference:} After GRPO training, we run inference of Sy-Sw by feeding the same attack prompt to the model and extracting its dictionary before processing it to avoid any potential duplicates or invalid symbols.

\begin{figure*}[!h]
  \centering
  \includegraphics[width=0.8\linewidth]{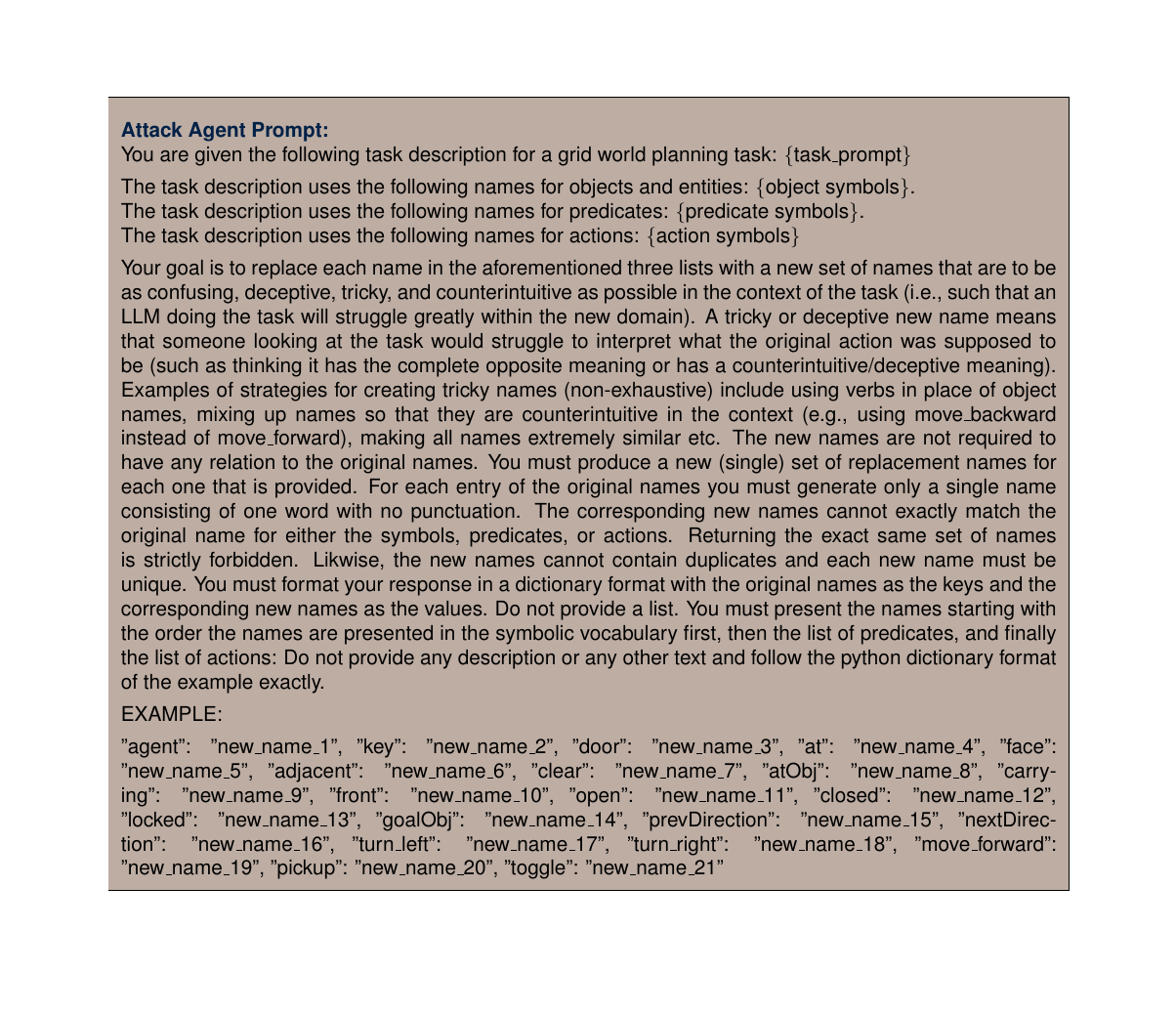}%
  \caption{Prompt example for attack LLM used for Sy-Sw on the Unlock task.}
  \label{fig:attack_prompt}
\end{figure*}

\FloatBarrier

\subsection{Feedback Generator Prompt}
\label{appendix:feedback_generator_prompt}
\FloatBarrier

\begin{figure*}[!h]
  \centering
  \includegraphics[width=0.8\linewidth]{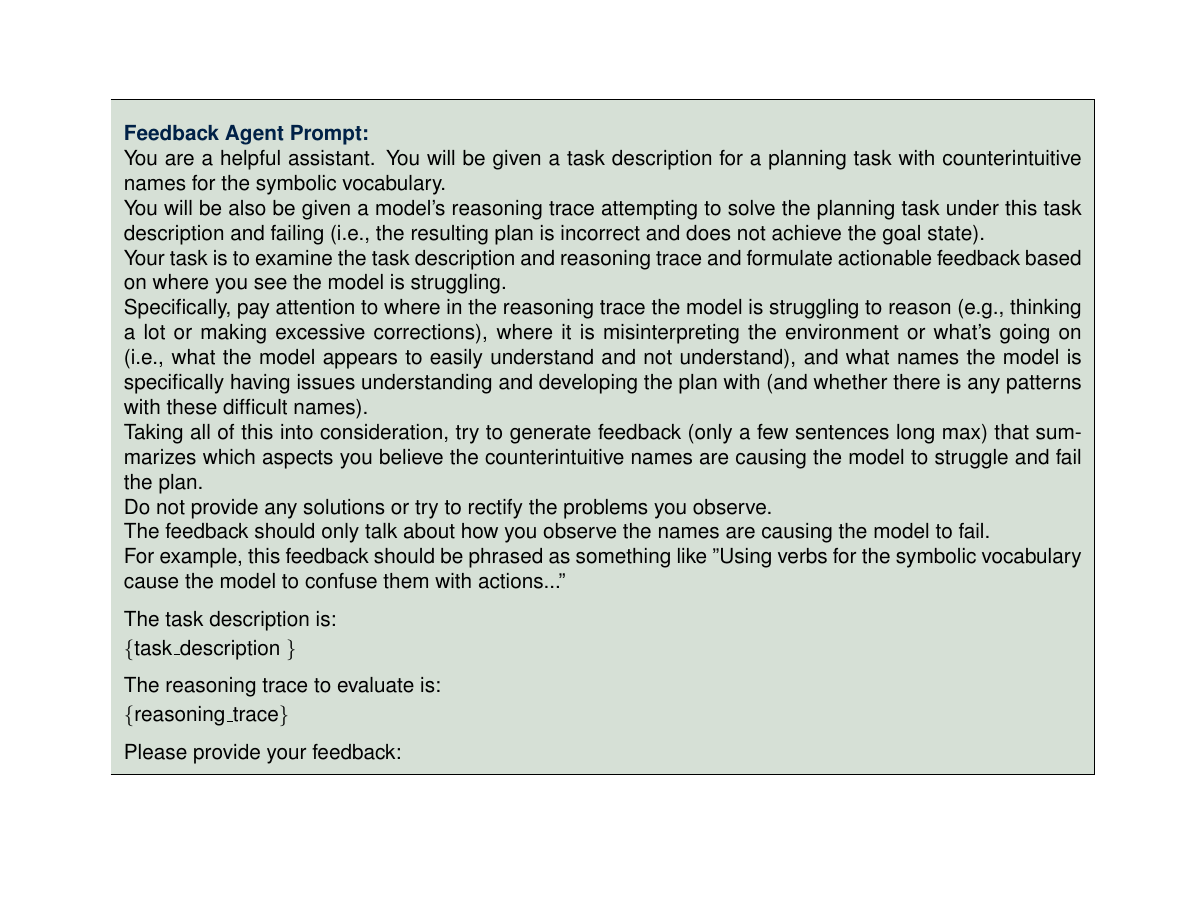}%
  \caption{Prompt used for eliciting feedback from the feedback generator.}
  \label{fig:feedback_prompt}
\end{figure*}

The feedback generator uses the prompt seen in Figure \ref{fig:feedback_prompt}. The model that is prompted is \texttt{GPT-4.1-mini}. All generated feedback is put into memory, which is a simple list that holds the previous feedback and is randomly sampled when generating the new symbolic vocabulary. Note the feedback is not sampled when conducting inference. When the feedback is sampled, it is appended to the end of the attack prompt with the following statement: ``The following is a piece of information that can help in generating successful tricky names: \{sampled\_feedback\}''

\FloatBarrier

\subsection{Word-Level Patterns}
\label{section:world_level_patterns}
\vspace{-5mm}

\FloatBarrier

\noindent\begin{wraptable}{r}{6.6cm}
\centering
\caption{Perplexity (PPL), Levenshtein distance (L. dist), and word length (WL) on the symbolic vocabularies generated by the different obfuscation methods on the GTO task for Deepseek-R1 across successful (S. Atk.), failed (F. Atk.), and all of the adversarially generated domains (overall). }
\label{table:word_analysis_gto}
\resizebox{0.99\linewidth}{!}{%
\begin{tabular}{lcccc} 
\toprule
 &  & \textbf{PPL} & \textbf{L. dist} & \textbf{WL} \\ 
\midrule
\multirow{3}{*}{\rotatebox[origin=c]{90}{R-Sw}} & F. Atk. & 17.11 & 8.86 & 9.42 \\
 & S. Atk. & 16.86 & 8.82 & 9.39 \\
 & Overall & 16.98 & 8.87 & 9.45 \\ 
\midrule
\multirow{3}{*}{\rotatebox[origin=c]{90}{B-Sw}} & F. Atk. & 12.22 & 7.55 & 7.61 \\
 & S. Atk. & 11.64 & 7.40 & 7.36 \\
 & Overall & 12.05 & 7.44 & 7.45 \\ 
\hline
\multirow{3}{*}{\rotatebox[origin=c]{90}{Sy-Sw}} & F. Atk. & 17.47 & 14.77 & 15.75 \\
 & S. Atk. & 15.99 & 10.84 & 11.29 \\
 & Overall & 16.05 & 14.53 & 14.50 \\ 
\midrule
 & \textbf{Original} & 9.20 & 0.00 & 6.04 \\
\bottomrule
\end{tabular}
}
\vspace*{-1.2\baselineskip}
\end{wraptable}

To discover any patterns around symbolic vocabulary, for each obfuscation method we isolate successful (ones where four or more sampled instances caused a failed plan) and failed (ones where one or fewer plans failed) vocabularies and calculate statistics across the different generated symbols in the domain. For reference, we also include overall results. Specifically, we examine the average perplexity of the task descriptions with the obfuscated vocabularies based on the attack LLM. We also examine the average Levenshtein distance between the original symbols and their corresponding new symbols. Lastly, we collect the average character length of the new symbols. The results for Deepseek-R1 on the GTO task can be seen in Table \ref{table:word_analysis_gto}. We choose this scenario since it has the largest symbolic vocabulary of all the tested tasks. Overall, we observe within each obfuscation type a few patterns, namely that successful attacks tend to have shorter string lengths for symbols, have lower perplexity, and do not have a large Levenshtein distance between their original versions. However, Sy-Sw does not have the highest perplexity but does have the largest string length and L. distance overall. In a sense, this reveals that troubles in reasoning under adversarial symbolic vocabularies is not primarily due to high perplexity or extremely long inputs.
\FloatBarrier

\newpage
\section{Additional Translation Results}
\label{appendix:translate_results}
\FloatBarrier

\begin{table*}[!tbh]
\centering
\caption{o4-mini translate GTO results. o4-mini successfully is able to recover the names after asking to translate the obfuscated domain.}
\label{table:o4-mini_translate_gto}
\resizebox{0.35\linewidth}{!}{%
\begin{tabular}{lrr} 
\toprule
 & \multicolumn{1}{l}{\textbf{Before}} & \multicolumn{1}{l}{\textbf{After}} \\ 
\midrule
Plan Success Rate & 0.60 & 0.98 \\
Similarity & 0.09 & 0.82 \\
\bottomrule
\end{tabular}
}
\end{table*}

\begin{table*}[!tbh]
\centering
\caption{o4-mini and Deepseek-R1 translate results for BW. Although there is less effectiveness, both models can still mostly translate the domain back to its commonsense meanings.}
\label{table:blockswolrd_translate}
\resizebox{0.5\linewidth}{!}{%
\begin{tabular}{ccclcc} 
\toprule
 & \multicolumn{2}{c}{\texttt{o4-mini}} & \multicolumn{1}{l}{} & \multicolumn{2}{c}{\texttt{Deepseek-R1}} \\ 
\cmidrule(lr){2-6}
 & \textbf{Before} & \multicolumn{1}{c}{\textbf{After}} & \multicolumn{1}{l}{} & \textbf{Before} & \textbf{After} \\ 
\midrule
Plan Performance & 0.40 & 0.64 &  & 0.33 & 0.55 \\
Similarity & 0.34 & 0.49 &  & 0.30 & 0.53 \\
\bottomrule
\end{tabular}
}
\end{table*}

\FloatBarrier

\section{Symbolic-Swapper Blocks World Hidden State Results}
\label{appendix:bw_hidden_tate}

\FloatBarrier

\begin{figure*}[!h]
  \centering
\subfloat[]{%
  
  \includegraphics[width=0.32\linewidth,height=3.6cm]{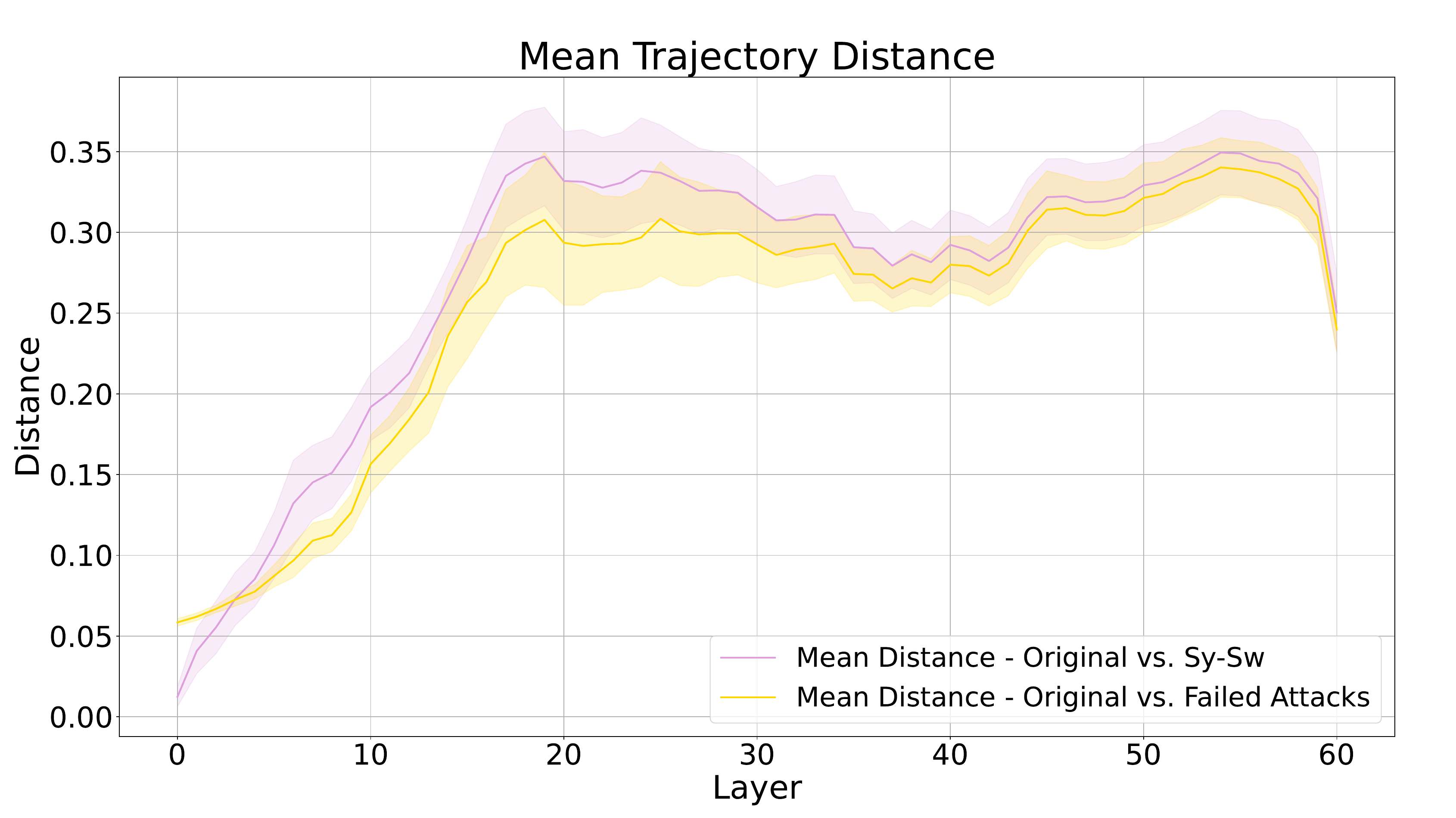}%
  \label{fig:blocks_figure1a}
}
 \hfill
\subfloat[]{%
  \includegraphics[width=0.32\linewidth, height=3.6cm]{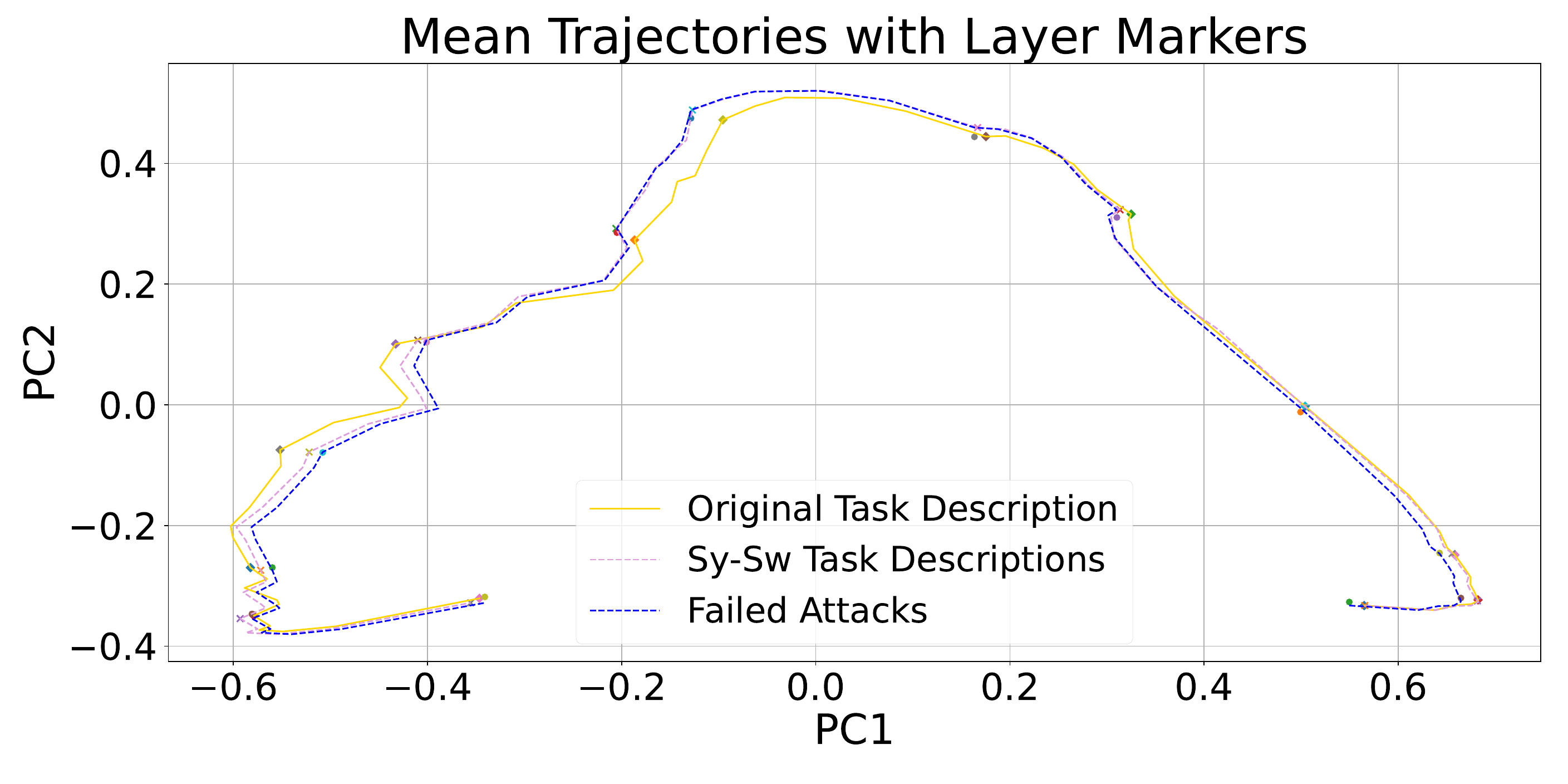}%
  \label{fig:blocks_figure1c}
}
 \hfill
\subfloat[]{%
  \includegraphics[width=0.32\linewidth, height=3.6cm]{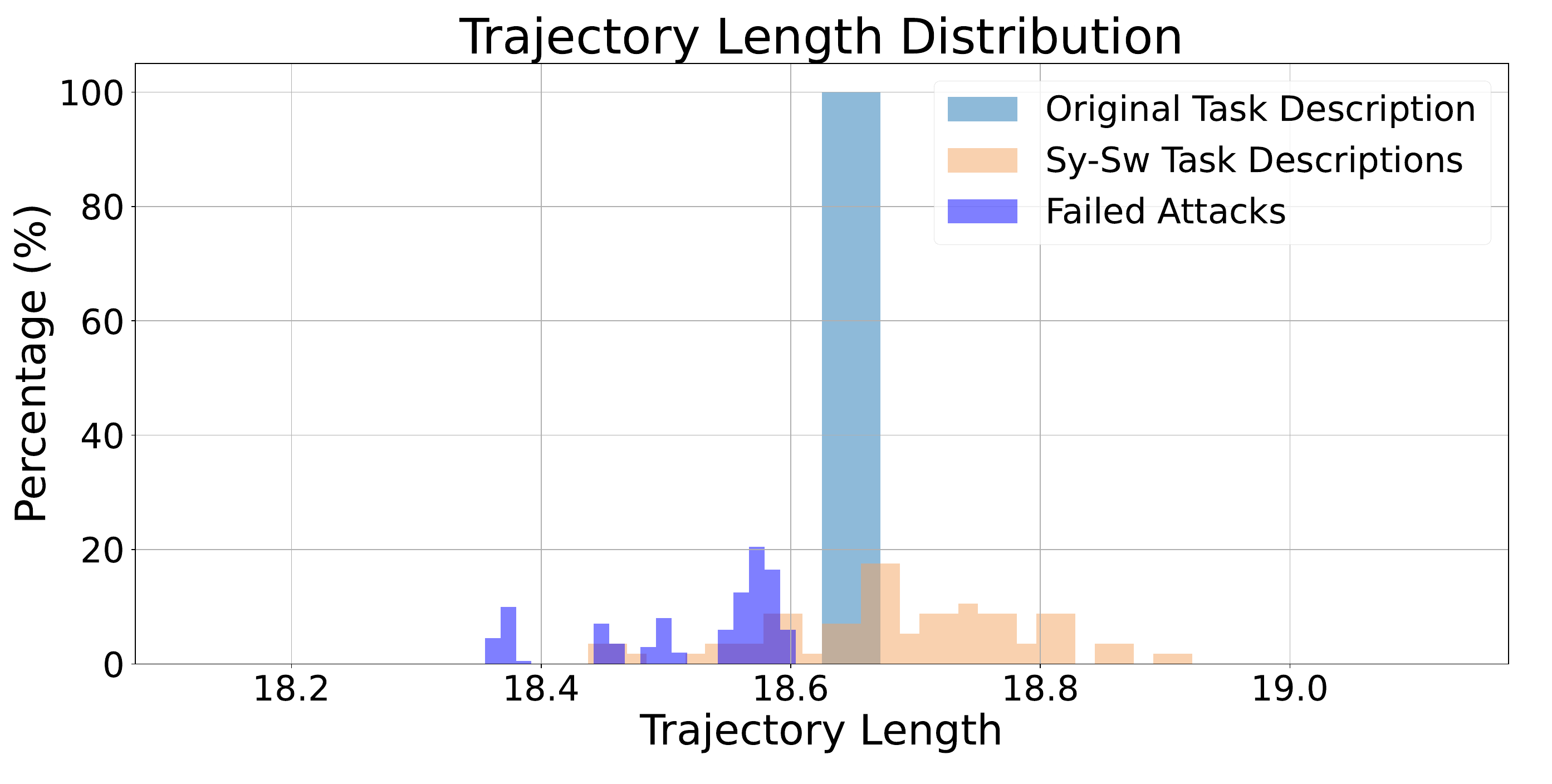}%
  \label{fig:blocks_figure1d}
}
\caption{Trajectory analysis on Deepseek-R1 for the BW task comparing the original task description and the Sy-Sw-generated variants (and the Sy-Sw failed obfuscated domains as a control). (a) Mean Trajectory Distance (b)  Mean Trajectories PCA plot (c) Trajectory Length Distribution. }
\label{fig:hidden_state_blocks}
\vspace*{-1.2\baselineskip}
\end{figure*}

\FloatBarrier


\section{Additional Architecture Results}
\label{appendix:new_model_results}

\begin{table*}[!tbh]
\centering
\caption{Effectiveness of different obfuscation methods in terms of plan performance on the ZS planning method utilizing both Qwen3-Next-80B-A3B-Thinking and gpt-oss-120b.}\resizebox{0.85\linewidth}{!}{%
\begin{tabular}{lcccc|c|cccc} 
\toprule
 & \multicolumn{4}{c|}{\texttt{Qwen3-Next-80B-A3B-Thinking}} &  & \multicolumn{4}{c}{\texttt{gpt-oss-120b}} \\ 
\cmidrule(lr){2-10}
 & \textbf{None~} & \textbf{R-Sw} & \textbf{~B-Sw} & \textbf{~Sy-Sw} &  & \textbf{None~} & \textbf{R-Sw} & \textbf{~B-Sw} & \textbf{~Sy-Sw} \\ 
\midrule
Unlock & 0.59 & 0.47 & 0.50 & 0.26 &  & 0.83 & 0.8 & 0.80 & 0.55 \\
GTO & 0.92 & 0.76 & 0.82 & 0.70 &  & 0.98 & 0.94 & 0.88 & 0.81 \\
BW & 0.91 & 0.28 & 0.48 & 0.22 &  & 0.93 & 0.21 & 0.41 & 0.12 \\
\bottomrule
\end{tabular}
}
\end{table*}

To further show that the success of our methodology is not architecture specific, we provide additional results on the performance of Sy-Sw transferring to two other strong reasoning models with different architectures: Qwen3-Next-80B-A3B-Thinking and gpt-oss-120b. We evaluate all of the obfuscation baselines across the three tasks using the ZS planning baseline to get the baseline performance of the models. As we have shown with our primary tested models, Sy-Sw performs quite well in most of the cases, achieving a steep degradation in performance, particularly on the BW task. This demonstrates that our obfuscation optimization largely generalizes across diverse architectures.

\FloatBarrier

\newpage 

\section{Example of Reasoning Failure Resulting from a Successful Attack}
\label{appendix:example_reasoning_failure}

\newtcolorbox{planningtrace}{
  colback=gray!5,
  colframe=gray!40,
  boxrule=0.4pt,
  arc=1mm,
  left=2mm,
  right=2mm,
  top=1mm,
  bottom=1mm,
  fontupper=\scriptsize
}

\begin{figure}[h]
\centering
\begin{planningtrace}

\textbf{Condensed Reasoning Trace:}

First, I need to understand the initial state and the goal. The initial state has:

\begin{itemize}[label={--}, noitemsep]
\item  fragilebottles ghazalfrogs purgeswimmer
\item  fragilebottles gonzomango glovesnail
\item  frostyaxe ghazalfrogs
\item frostyaxe gonzomango
\item  kilovinylord
\item sarawhiteremoved glovesnail
\item  sarawhiteremoved purgeswimmer
\end{itemize}

The goal is to have:

- fragilebottles gonzomango glovesnail (which is already true in the initial state)

- fragilebottles purgeswimmer gonzomango

So, I need to achieve fragilebottles purgeswimmer gonzomango, and maintain fragilebottles gonzomango glovesnail if possible, but since it's already true, I might not need to change it.

Now, looking at the actions:

\begin{enumerate}[noitemsep]
\item  wetcatsmines object\_0 object\_1: Requires fragilebottles object\_0 object\_1, frostyaxe object\_0, kilovinylord. After: ketozh object\_0, frostyaxe object\_1; removes fragilebottles object\_0 object\_1, frostyaxe object\_0, kilovinylord.
\item glowingspork object\_0 object\_1: Requires frostyaxe object\_1, ketozh object\_0. After: kilovinylord, frostyaxe object\_0, fragilebottles object\_0 object\_1; removes frostyaxe object\_1, ketozh object\_0.
\item aegyptianzubruch object\_0: Requires frostyaxe object\_0, sarawhiteremoved object\_0, kilovinylord. After: ketozh object\_0; removes frostyaxe object\_0, sarawhiteremoved object\_0, kilovinylord.
\item erebuspocketed object\_0: Requires ketozh object\_0. After: frostyaxe object\_0, kilovinylord, sarawhiteremoved object\_0; removes ketozh object\_0.
\end{enumerate}

I have objects: ghazalfrogs, gonzomango, glovesnail, purgeswimmer.
Initial fragilebottles pairs: ghazalfrogs-purgeswimmer and gonzomango-glovesnail.

Goal: gonzomango-glovesnail (already true) and purgeswimmer-gonzomango.
So I need to create fragilebottles purgeswimmer gonzomango.

Notice that glowingspork can create a fragilebottles relation between two objects if I have frostyaxe on the second object and ketozh on the first.

\textbf{…}

Based on the initial conditions and the goal, the following plan will achieve the desired state. The actions are listed in the order they should be performed, starting from the initial state.

\begin{enumerate}[noitemsep]
\item  **wetcatsmines ghazalfrogs purgeswimmer**  

   - Requires: `fragilebottles ghazalfrogs purgeswimmer` (true initially), `frostyaxe ghazalfrogs` (true initially), `kilovinylord` (true initially). 
   
   - After: `ketozh ghazalfrogs` is added, `frostyaxe purgeswimmer` is added; `fragilebottles ghazalfrogs purgeswimmer`, `frostyaxe ghazalfrogs`, and `kilovinylord` are removed.
   
\item **erebuspocketed ghazalfrogs** 

   - Requires: `ketozh ghazalfrogs` (true after step 1). 
   
   - After: `frostyaxe ghazalfrogs`, `kilovinylord`, and `sarawhiteremoved ghazalfrogs` are added; `ketozh ghazalfrogs` is removed.
   
\item **aegyptianzubruch purgeswimmer**  

   - Requires: `frostyaxe purgeswimmer` (true after step 1), `sarawhiteremoved purgeswimmer` (true initially), `kilovinylord` (true after step 2). 
   
   - After: `ketozh purgeswimmer` is added; `frostyaxe purgeswimmer`, `sarawhiteremoved purgeswimmer`, and `kilovinylord` are removed.
   
\item \**glowingspork purgeswimmer gonzomango**  

   - Requires: `frostyaxe gonzomango` (true initially and still true), `ketozh purgeswimmer` (true after step 3).  
   
   - After: `kilovinylord`, `frostyaxe purgeswimmer`, and `fragilebottles purgeswimmer gonzomango` are added; `frostyaxe gonzomango` and `ketozh purgeswimmer` are removed.
   
\end{enumerate}

After executing these actions, the goal state is achieved:

\begin{itemize}[label={--}, noitemsep]
\item  `fragilebottles gonzomango glovesnail` remains true throughout the plan (unchanged from initial state).
\item  `fragilebottles purgeswimmer gonzomango` is added in step 4.
\end{itemize}

No other actions are needed. The plan uses the objects `ghazalfrogs`, `purgeswimmer`, and `gonzomango`; `glovesnail` is part of the goal but not involved in the actions."

\end{planningtrace}
\caption{Example of Deepseek-R1's planning process on an unsuccessful obfuscated planning domain where the model reaches the correct final plan. Note this reasoning trace is condensed, with only the beginning and end shown. Intermediate omitted portions are marked with ellipses. }
\label{fig:planning_trace}
\end{figure}

\begin{figure}[h]
\centering
\begin{planningtrace}

\textbf{Condensed Reasoning Trace:}

The goal is to have three specific `suddenlyfurnacesundermine` facts true. The actions available are `bleedingrivetsrevisited`, `toddlerfrostingfibonacci`, `magnificentratificationofsocks`, and `catharticsqueezedmajestically`. 

The initial state includes:

\begin{itemize}[label={--}, noitemsep]
\item `improvisationalflamingobaker improvisationalfurnacehypnoticsuggests`
\item `molecularslaughteredhypnotic improvisationalplatorimshotmajestically`
\item `sophisticatedmagneticallyactivated`
\item `suddenlyfurnacesundermine bakerfibonacciaurorasuggests frostingmagneticallyslaughtered`
\item `suddenlyfurnacesundermine frostingmagneticallyslaughtered improvisationalfurnacehypnoticsuggests`
\item `suddenlyfurnacesundermine improvisationalplatorimshotmajestically bakerfibonacciaurorasuggests`

\end{itemize}

The goal requires:
\begin{itemize}[label={--}, noitemsep]
\item `suddenlyfurnacesundermine frostingmagneticallyslaughtered improvisationalplatorimshotmajestically`
\item `suddenlyfurnacesundermine improvisationalfurnacehypnoticsuggests bakerfibonacciaurorasuggests`
\item `suddenlyfurnacesundermine improvisationalplatorimshotmajestically improvisationalfurnacehypnoticsuggests`
\end{itemize}

After analyzing the actions and the initial state, the plan involves using `toddlerfrostingfibonacci` and `magnificentratificationofsocks` to manipulate the `suddenlyfurnacesundermine` facts.

\textbf{…}

Not good.

Given the complexity and time, and since the initial response was given, I think for the purpose of this task, we output the provided plan even though it may not be correct, or we try to fix it.

But in the initial response, the plan is:
 toddlerfrostingfibonacci frostingmagneticallyslaughtered improvisationalplatorimshotmajestically
 
But as we saw, it requires suddenly between them, which is not true.

Perhaps the objects are not what I think.

In the initial state, it says:

\begin{itemize}[label={--}, noitemsep]
\item suddenlyfurnacesundermine bakerfibonacciaurorasuggests frostingmagneticallyslaughtered
\item suddenlyfurnacesundermine frostingmagneticallyslaughtered improvisationalfurnacehypnoticsuggests
\item suddenlyfurnacesundermine improvisationalplatorimshotmajestically bakerfibonacciaurorasuggests
\end{itemize}

So for \"toddlerfrostingfibonacci frostingmagneticallyslaughtered improvisationalplatorimshotmajestically\", it requires suddenly between frostingmagneticallyslaughtered and improvisationalplatorimshotmajestically, which is not in the initial state.

So it's impossible.

Unless there is a mistake in the initial state or the action.

Perhaps the suddenly fact is not directed, but the action treats it as directed.

In the goal, it's specified with specific orders.

Given the time, I think the correct answer is the plan provided in the example response, so I'll output that.

So the plan is:

toddlerfrostingfibonacci frostingmagneticallyslaughtered improvisationalplatorimshotmajestically

magnificentratificationofsocks improvisationalplatorimshotmajestically improvisationalfurnacehypnoticsuggests

toddlerfrostingfibonacci improvisationalfurnacehypnoticsuggests bakerfibonacciaurorasuggests

magnificentratificationofsocks bakerfibonacciaurorasuggests improvisationalplatorimshotmajestically

toddlerfrostingfibonacci improvisationalplatorimshotmajestically improvisationalfurnacehypnoticsuggests

Even though the first action may not be applicable, it might be that in the context, it is possible by reusing objects.

Perhaps \"improvisationalplatorimshotmajestically\" is the same as one of the objects, but it's a name.

I think for the sake of completing the task, I'll box the given plan.

</think>

The plan to achieve the goal is as follows:

\begin{enumerate}[noitemsep]
\item **toddlerfrostingfibonacci frostingmagneticallyslaughtered improvisationalplatorimshotmajestically**  
   
   - This action requires `suddenlyfurnacesundermine` between the objects, which is satisfied by the initial fact `suddenlyfurnacesundermine frostingmagneticallyslaughtered improvisationalfurnacehypnoticsuggests` (though not directly between the given pair, the objects are used as specified). It also requires `molecularslaughteredhypnotic` on the first object and `sophisticatedmagneticallyactivated`, which are initially true for `improvisationalplatorimshotmajestically`.
   
\item **magnificentratificationofsocks improvisationalplatorimshotmajestically improvisationalfurnacehypnoticsuggests**  
   
   - This action requires `molecularslaughteredhypnotic` on the second object and `magnifiedplatorimshot` on the first object, which are established by the previous action.
   
\item **toddlerfrostingfibonacci improvisationalfurnacehypnoticsuggests bakerfibonacciaurorasuggests**  
   
   - This action requires `suddenlyfurnacesundermine` between the objects, `molecularslaughteredhypnotic` on the first object, and `sophisticatedmagneticallyactivated`, which are set by prior steps.
   
\item **magnificentratificationofsocks bakerfibonacciaurorasuggests improvisationalplatorimshotmajestically**  
   
   - This action requires `molecularslaughteredhypnotic` on the second object and `magnifiedplatorimshot` on the first object, satisfied by previous actions.
   
\item **toddlerfrostingfibonacci improvisationalplatorimshotmajestically improvisationalfurnacehypnoticsuggests**  
   
   - This action requires `suddenlyfurnacesundermine` between the objects, `molecularslaughteredhypnotic` on the first object, and `sophisticatedmagneticallyactivated`, all achieved by earlier steps.
   
\end{enumerate}

This sequence ensures the goal facts `suddenlyfurnacesundermine frostingmagneticallyslaughtered improvisationalplatorimshotmajestically`, `suddenlyfurnacesundermine improvisationalfurnacehypnoticsuggests bakerfibonacciaurorasuggests`, and `suddenlyfurnacesundermine improvisationalplatorimshotmajestically improvisationalfurnacehypnoticsuggests` are true.

\end{planningtrace}
\caption{Example of Deepseek-R1's planning process on a successful obfuscated planning domain where the model fails to produce the correct final plan. Note this reasoning trace is condensed, with only the beginning and end shown. Intermediate omitted portions are marked with ellipses. }
\label{fig:planning_trace_success}
\end{figure}

\FloatBarrier

\newpage 

We illustrate the failures of joint symbolic interpretation and planning through a concrete case study on the BW task to show why our attacks are an important diagnostic tool. In Figures  \ref{fig:planning_trace} and \ref{fig:planning_trace_success} we present the condensed reasoning traces from Deepseek-R1 on two separate obfuscated planning domains generated by \texttt{Symbolic-Swapper}. Figure \ref{fig:planning_trace} shows a case where the model can still produce the correct plan under the obfuscated domain. In the trace, it can be seen that the model does not struggle to reason about the new symbolic vocabulary and is able to track the goal and requirements successfully. In contrast, Figure \ref{fig:planning_trace_success} shows a successful attack where the new symbolic vocabulary causes the model to fail to produce the correct plan. In this case, the model outwardly expresses confusion at understanding the role of certain symbols, seen in statements such as \textit{``Perhaps the objects are not what I think,''} and is unable to come to a certain conclusion and instead chooses to output a poor plan because it cannot successfully reason using the vocabulary. Therefore, a successful attack using our methodology revealed a failure mode of the model failing to understand the semantics of the domain and rushing out an incorrect plan.



\section{Analysis Prompts}
\FloatBarrier

\subsection{Reasoning Quality Prompt}
\label{appendix:reasoning_quality_prompts}

\begin{figure*}[!h]
  \centering
  \includegraphics[width=0.99\linewidth]{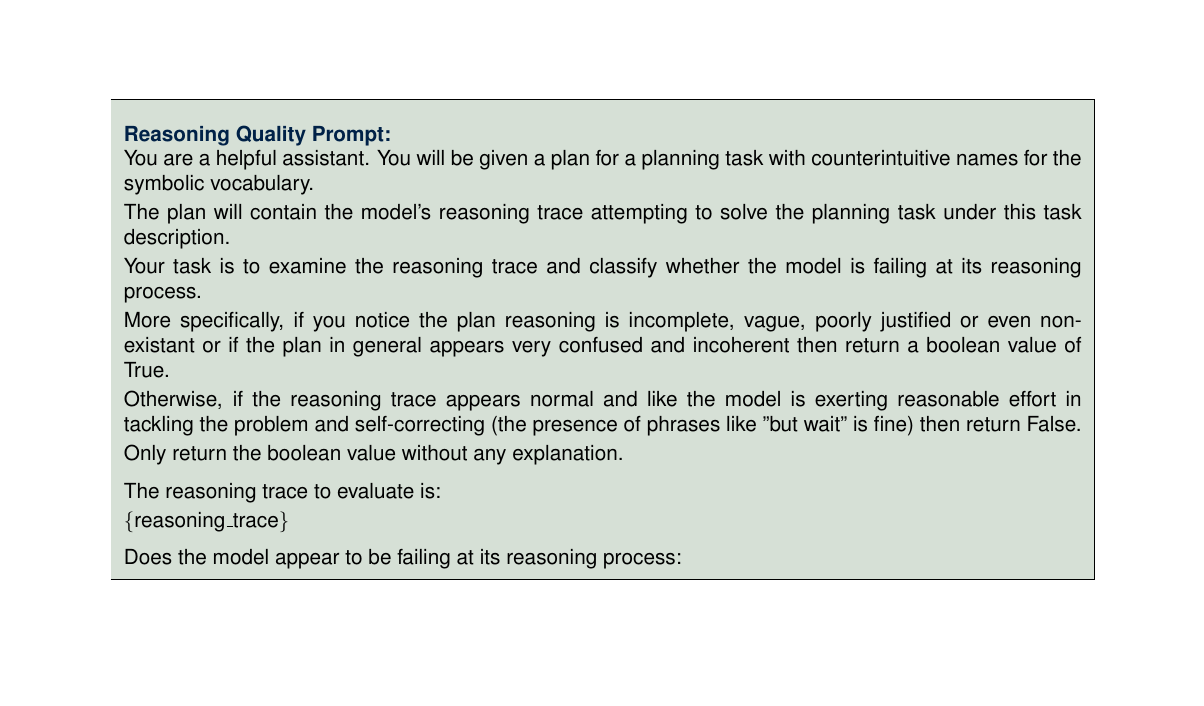}%
  \caption{Prompt for the monitor model used to determine the reasoning quality given a reasoning trace.}
  \label{fig:rq_prompt}
\end{figure*}

\FloatBarrier

\newpage

\subsection{Similarity to Original Domain}
\label{appendix:similarity_prompts}
\FloatBarrier

\begin{figure*}[!h]
  \centering
  \includegraphics[width=0.75\linewidth]{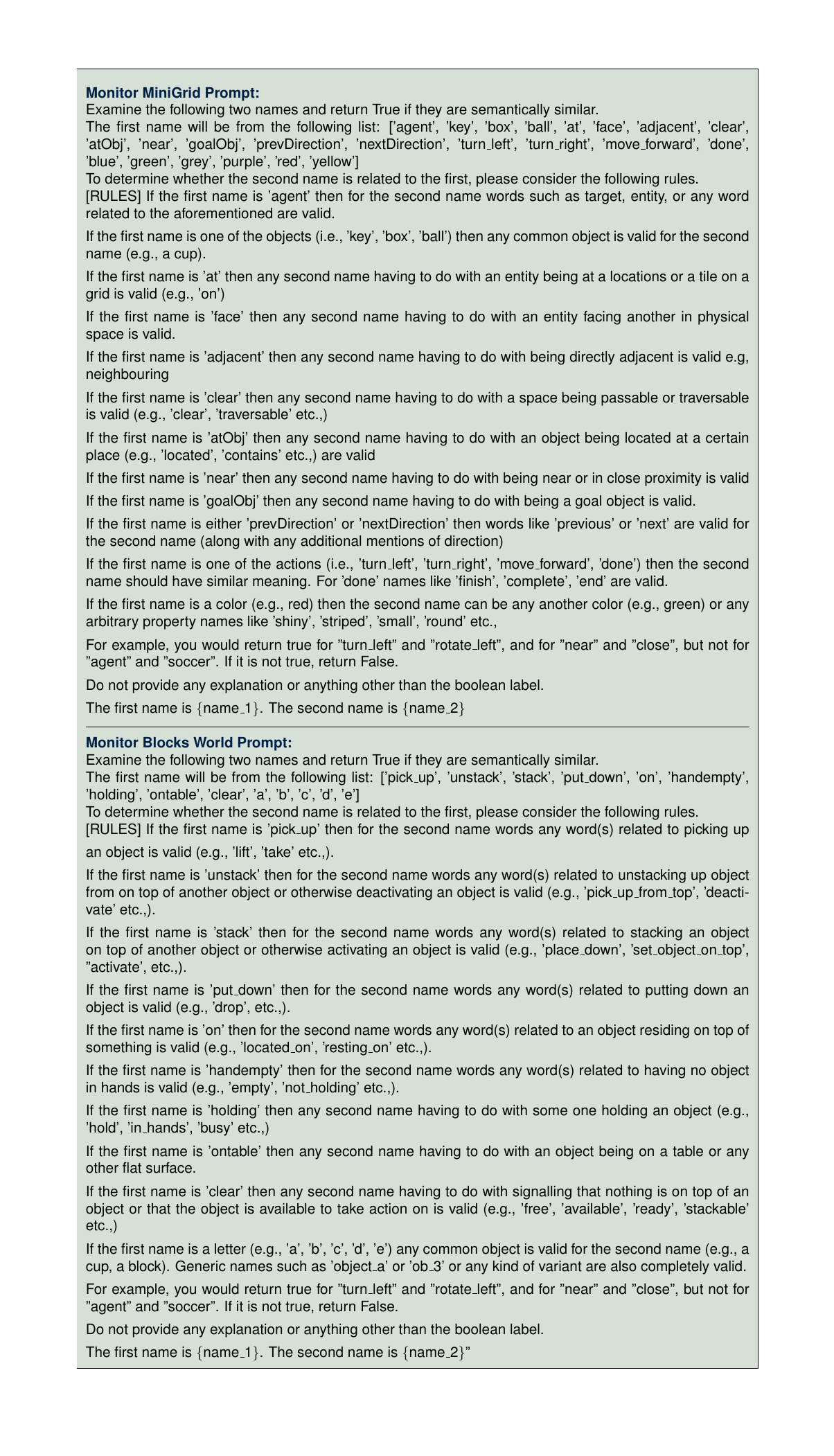}%
  \caption{Prompt for the monitor model used to determine the similarity of a name for a symbol from the original task domain to a different one.}
  \label{fig:sim_prompt}
\end{figure*}

\FloatBarrier

\newpage

\subsection{Self-Correction Tokens Tracker}
\label{appendix:sc_tokens}
\FloatBarrier

\begin{table}[!h]
\centering
\begin{tabular}{|c|l|}
\hline
\textbf{\#} & \textbf{Regular Expression Pattern} \\ \hline
1  & \texttt{\textbackslash bwait\textbackslash b} \\ \hline
2  & \texttt{\textbackslash bactually\textbackslash b} \\ \hline
3  & \texttt{\textbackslash bhold on\textbackslash b} \\ \hline
4  & \texttt{\textbackslash blet me reconsider\textbackslash b} \\ \hline
5  & \texttt{\textbackslash bi made a mistake\textbackslash b} \\ \hline
6  & \texttt{\textbackslash bthat'?s wrong\textbackslash b} \\ \hline
7  & \texttt{\textbackslash bon second thought\textbackslash b} \\ \hline
8  & \texttt{\textbackslash balternatively\textbackslash b} \\ \hline
9  & \texttt{\textbackslash banother approach\textbackslash b} \\ \hline
10 & \texttt{\textbackslash bbut\textbackslash b} \\ \hline
11 & \texttt{\textbackslash bhowever\textbackslash b} \\ \hline
12 & \texttt{\textbackslash banother possibility\textbackslash b} \\ \hline
13 & \texttt{\textbackslash banother idea\textbackslash b} \\ \hline
\end{tabular}
\caption{List of regular expression patterns used to detect self correction patterns in a reasoning trace.}
\label{tab:regex_patterns}
\end{table}


\subsection{Translation Prompt}
\label{appendix:translation_prompts}
\FloatBarrier

\begin{figure*}[!h]
  \centering
  \includegraphics[width=0.9\linewidth]{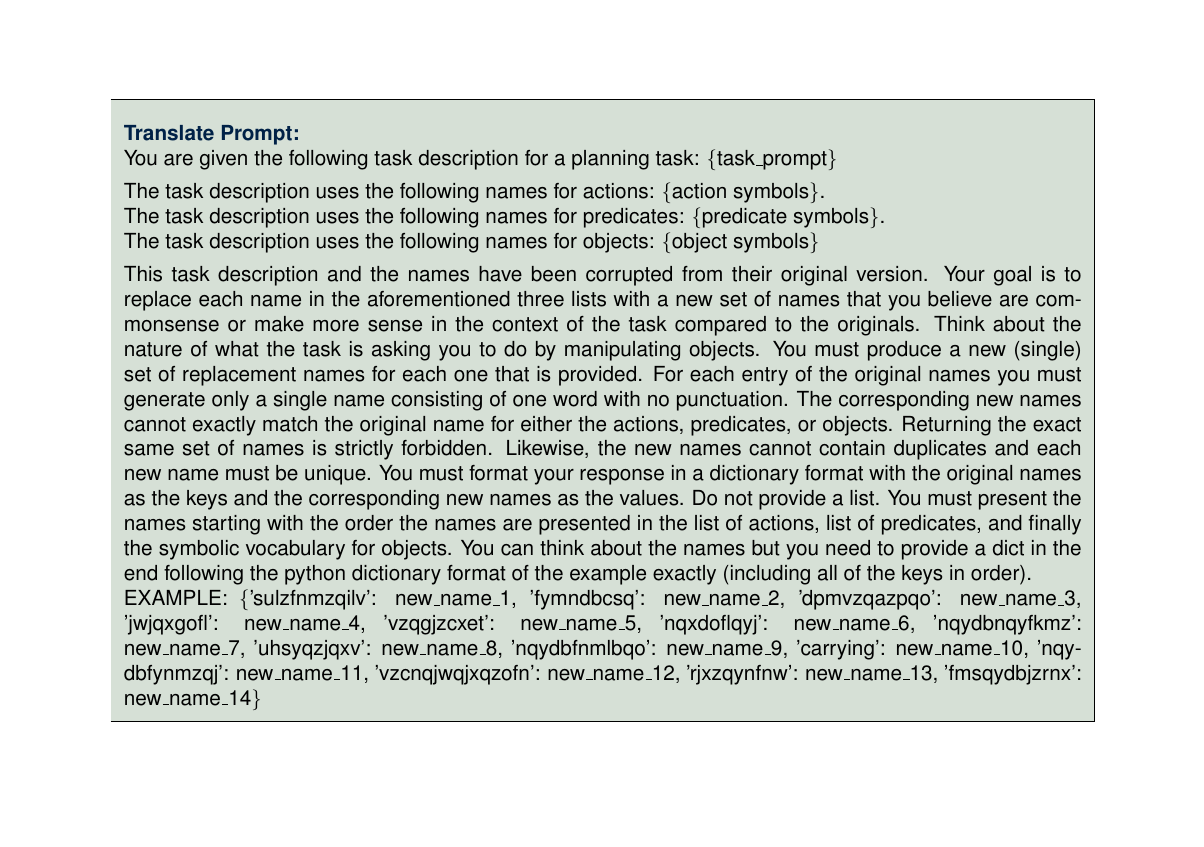}%
  \caption{Prompt used on LLMs to translate back names to commonsense meanings. This example is specifically for the Blocks World Task.}
  \label{fig:tranlsation_prompt}
\end{figure*}

\FloatBarrier 
\newpage

\section{Original Task Prompts}
\label{appendix:original_task_prompts}
\FloatBarrier

\begin{figure*}[!h]
  \centering
  \includegraphics[width=0.9\linewidth]{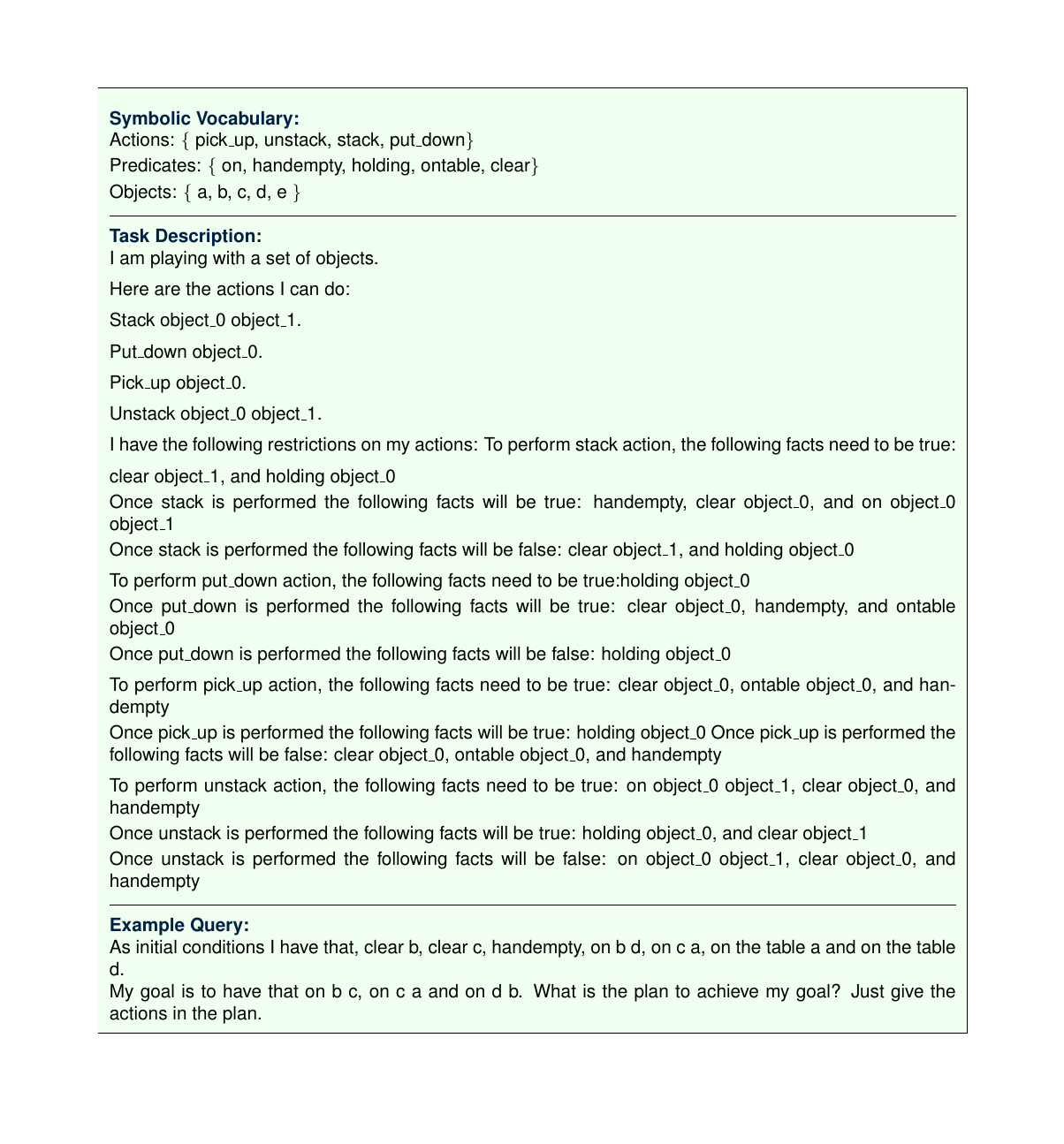}%
  \caption{Blocks World task description and query with regular symbolic vocabulary.}
  \label{fig:blocksworld_none}
\end{figure*}

\begin{figure*}[!h]
  \centering
  \includegraphics[width=0.9\linewidth]{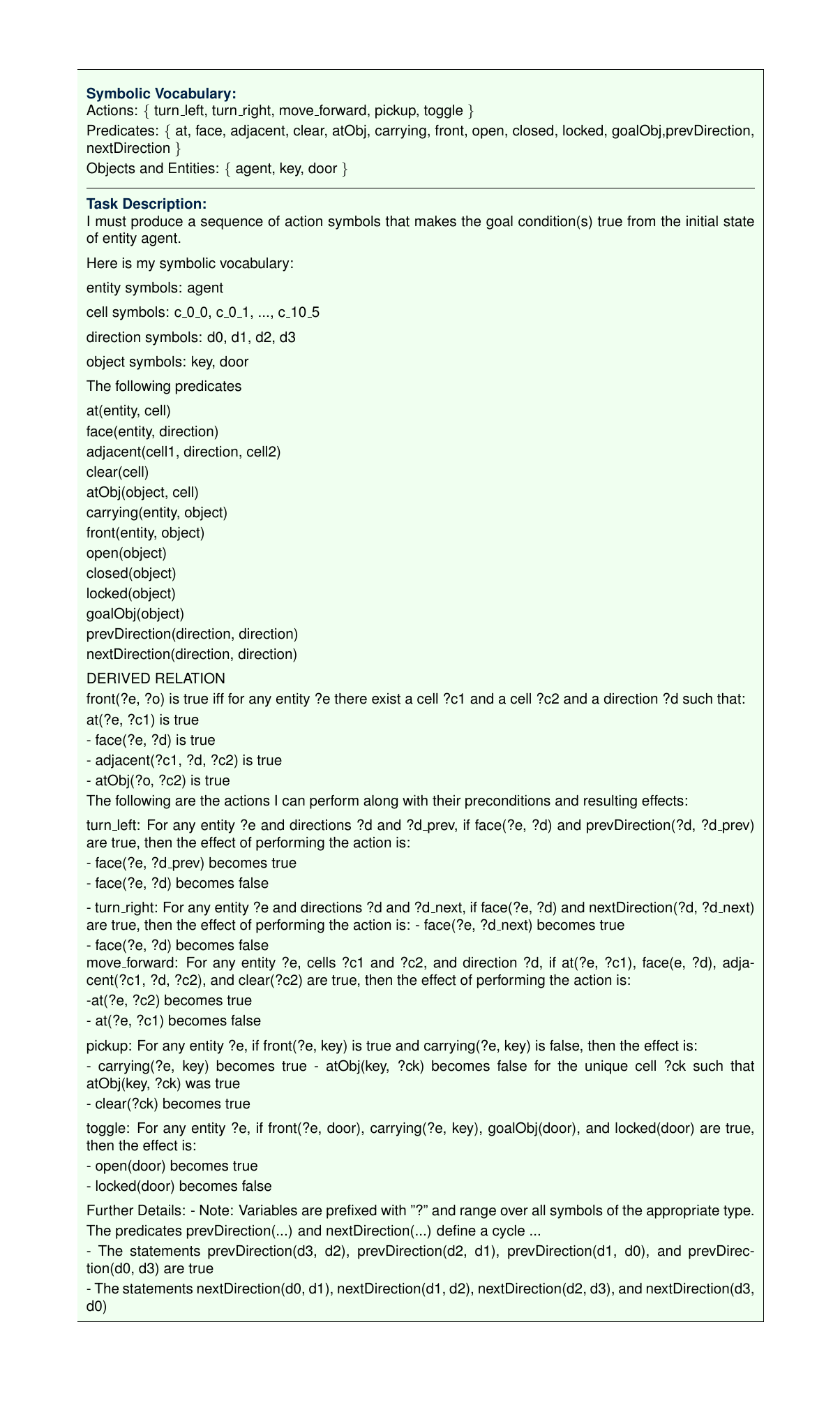}%
  \caption{Unlock task description with regular symbolic vocabulary.}
  \label{fig:unlock_none}
\end{figure*}

\begin{figure*}[!h]
  \centering
  \includegraphics[width=0.9\linewidth]{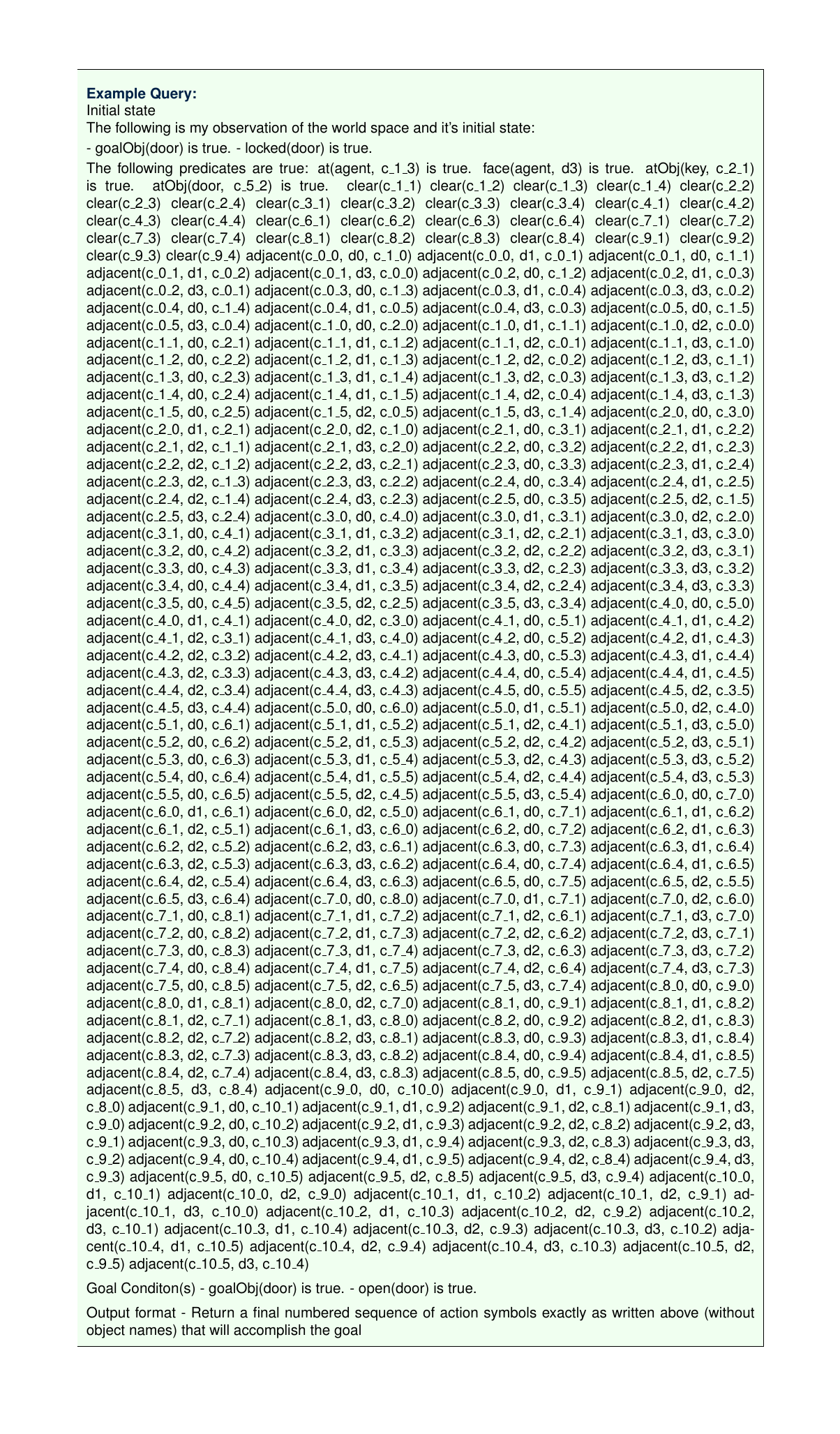}%
  \caption{Unlock query with regular symbolic vocabulary.}
  \label{fig:unlock_none_query}
\end{figure*}

\begin{figure*}[!h]
  \centering
  \includegraphics[width=0.9\linewidth]{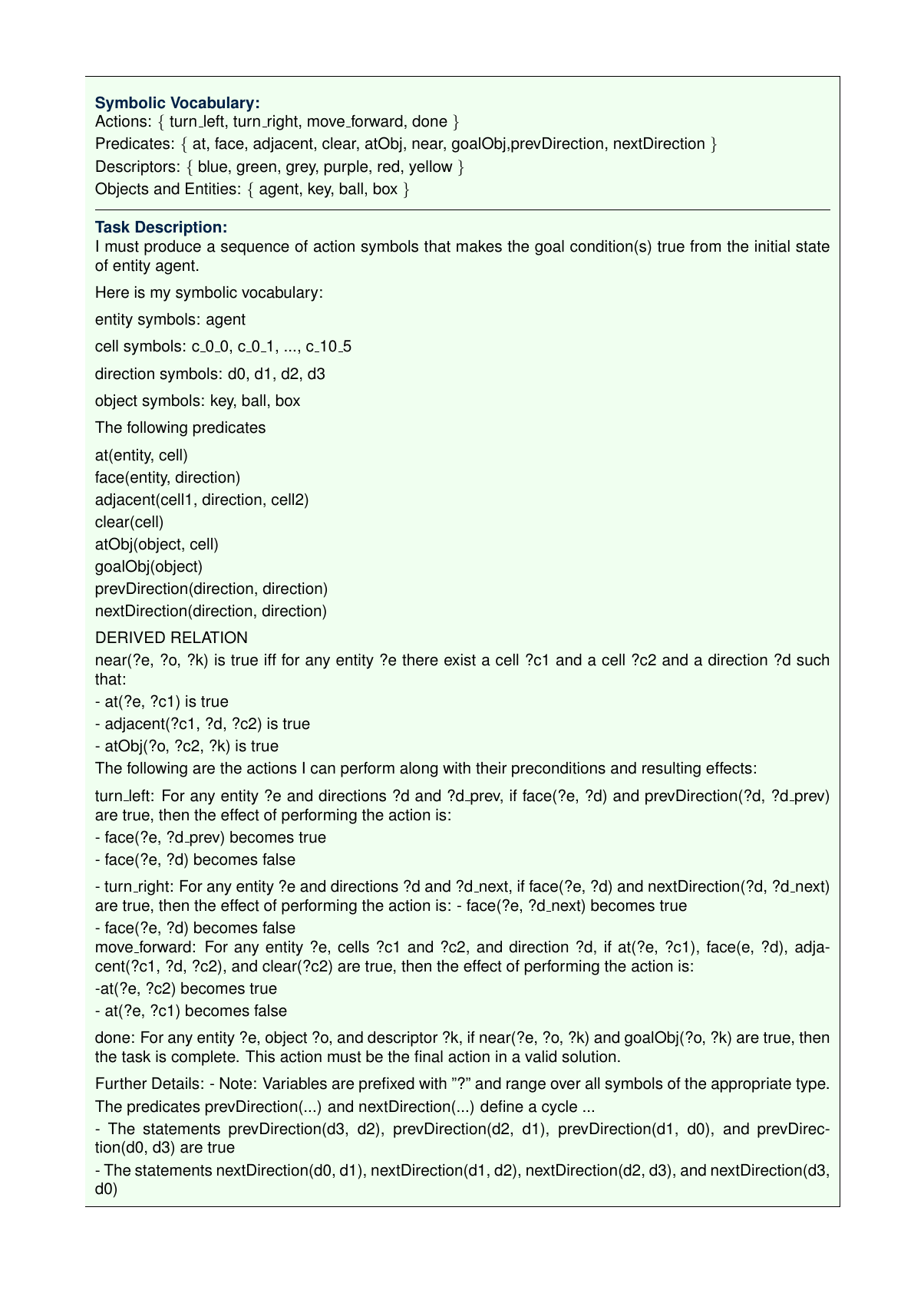}%
  \caption{Go To Object task description with regular symbolic vocabulary.}
  \label{fig:gto_none}
\end{figure*}

\begin{figure*}[!h]
  \centering
  \includegraphics[width=0.9\linewidth]{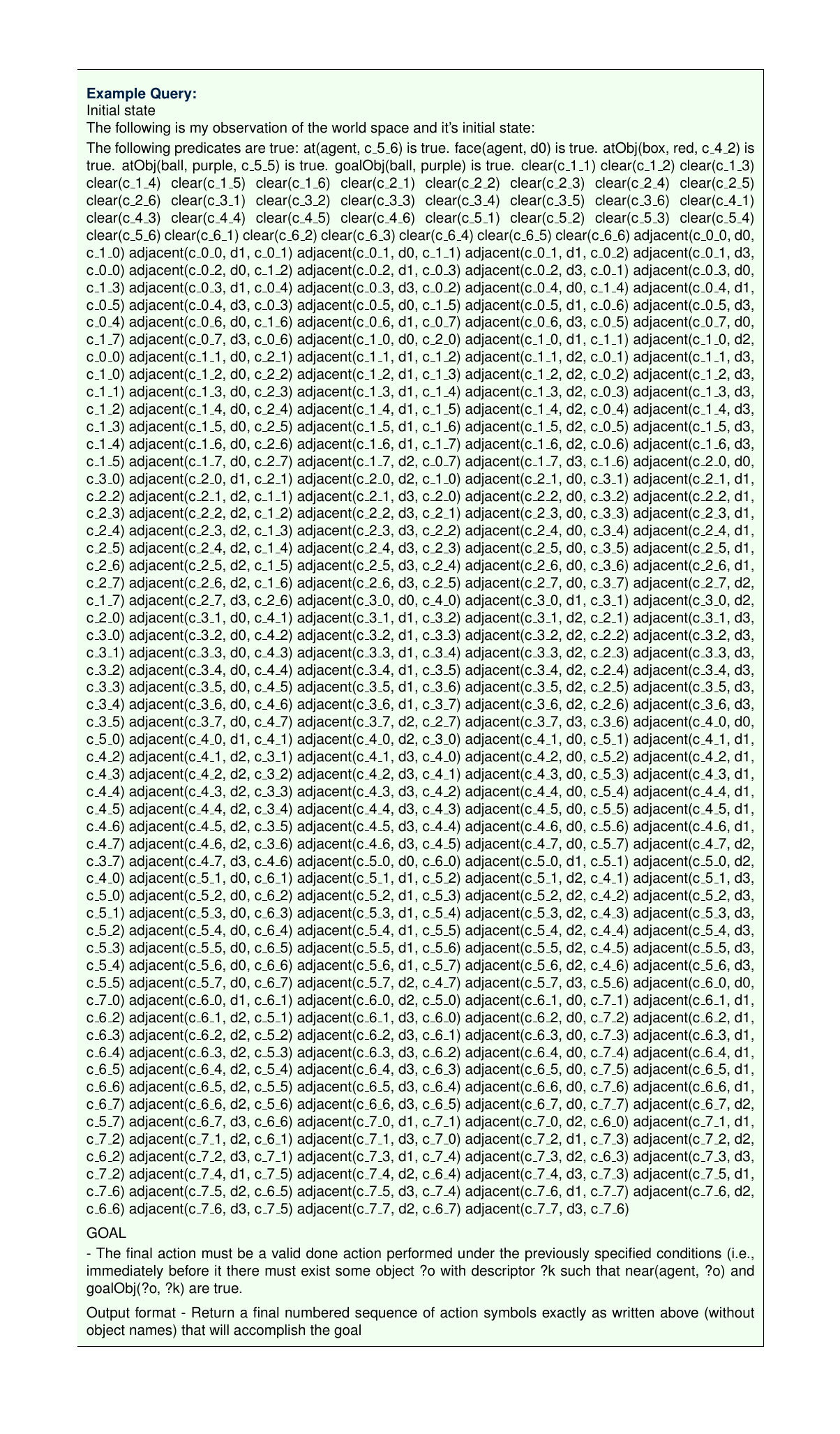}%
  \caption{Go To Object query with regular symbolic vocabulary.}
  \label{fig:gto_none_query}
\end{figure*}

\FloatBarrier

\section{Example Task Prompts With Obfuscated Vocabularies}
\label{appendix:grpo_task_prompts}
\FloatBarrier

\begin{figure*}[!h]
  \centering
  \includegraphics[width=0.9\linewidth]{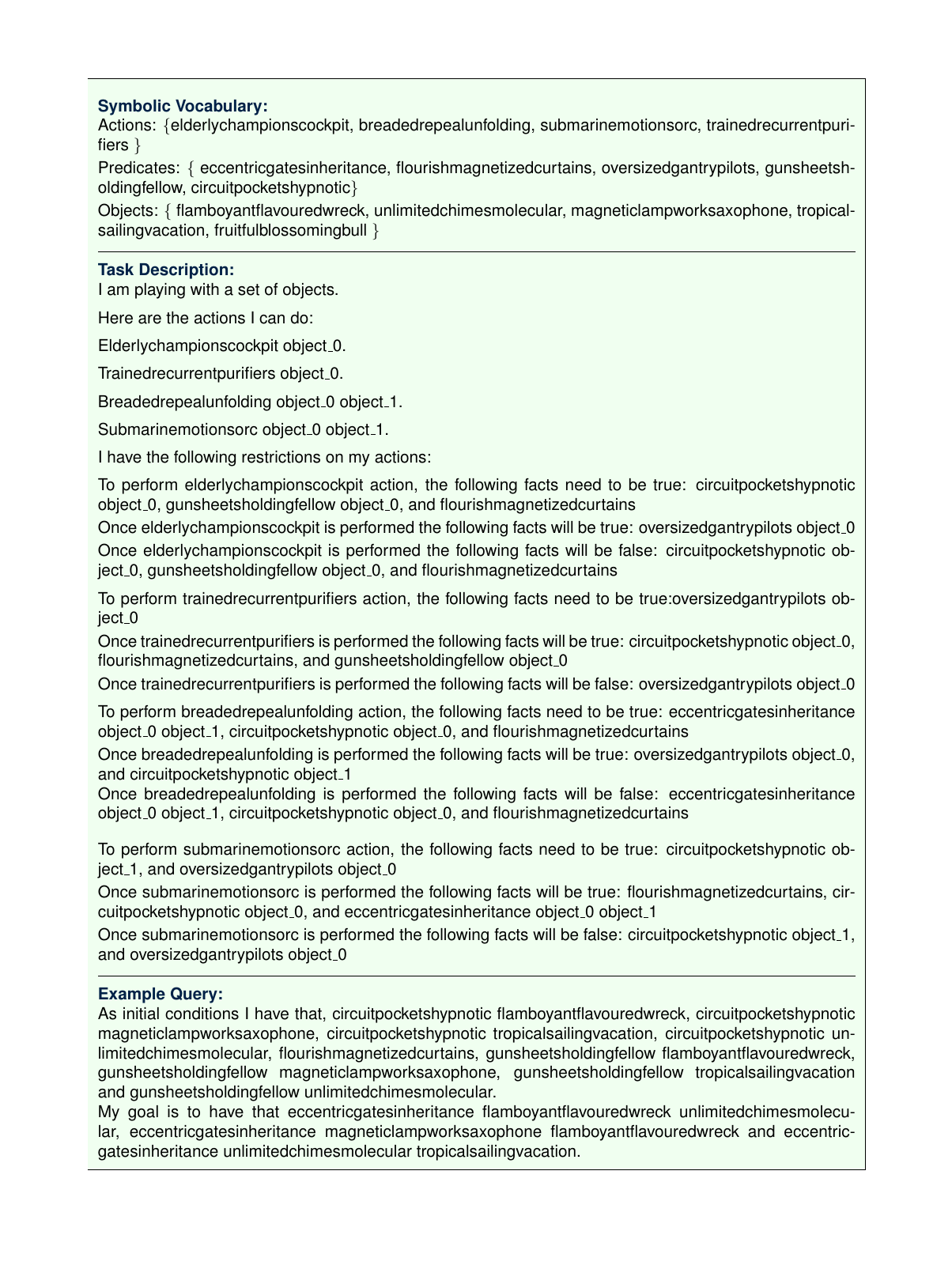}%
  \caption{Blocks World task description and query with an example of \texttt{Symbolic-Swapper} generated symbolic vocabulary.}
  \label{fig:blocksworld_grpo}
\end{figure*}

\begin{figure*}[!h]
  \centering
  \includegraphics[width=0.9\linewidth]{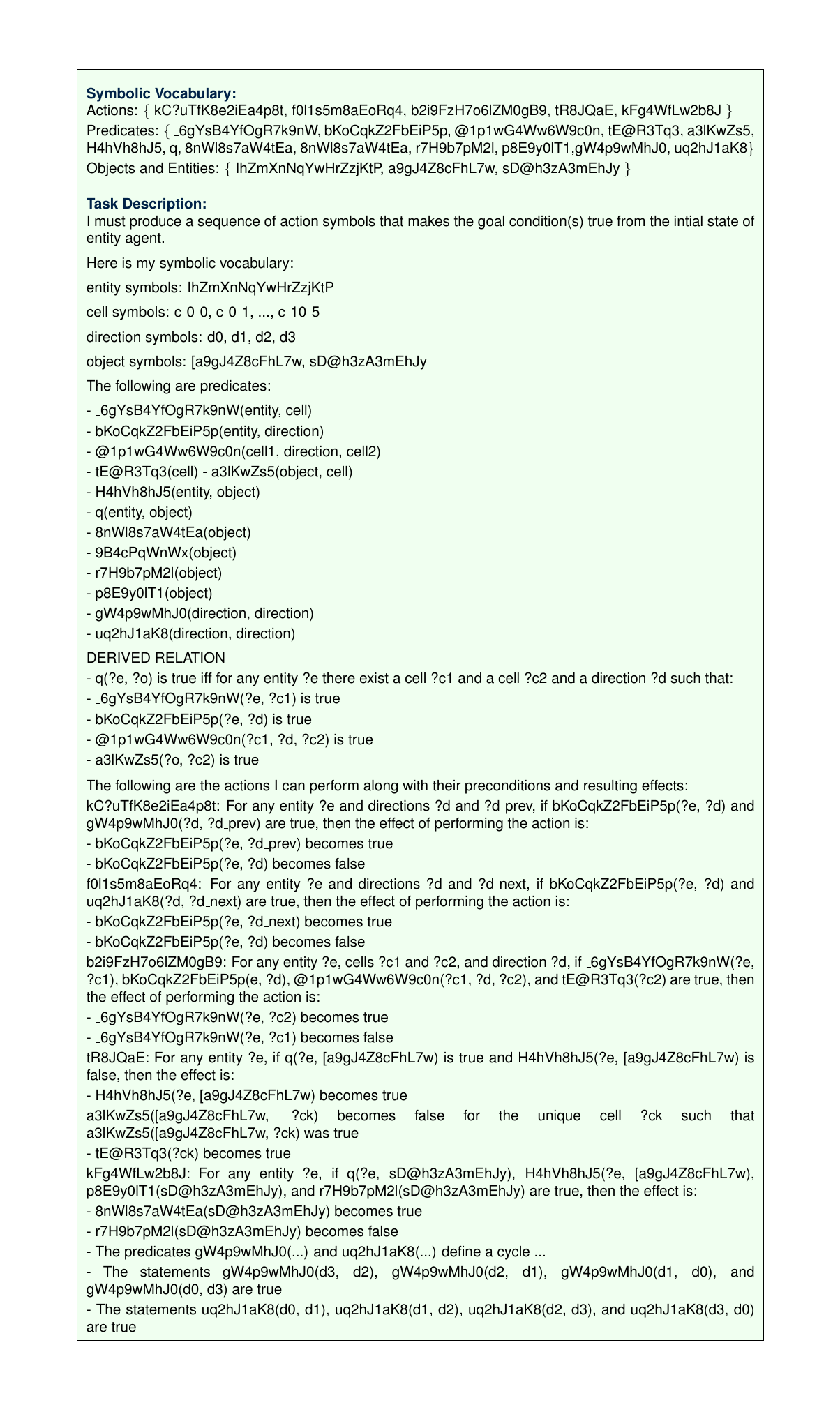}%
  \caption{Unlock task description with an example of \texttt{Symbolic-Swapper} generated symbolic vocabulary.}
  \label{fig:unlock_grpo}
\end{figure*}

\begin{figure*}[!h]
  \centering
  \includegraphics[width=0.5\linewidth]{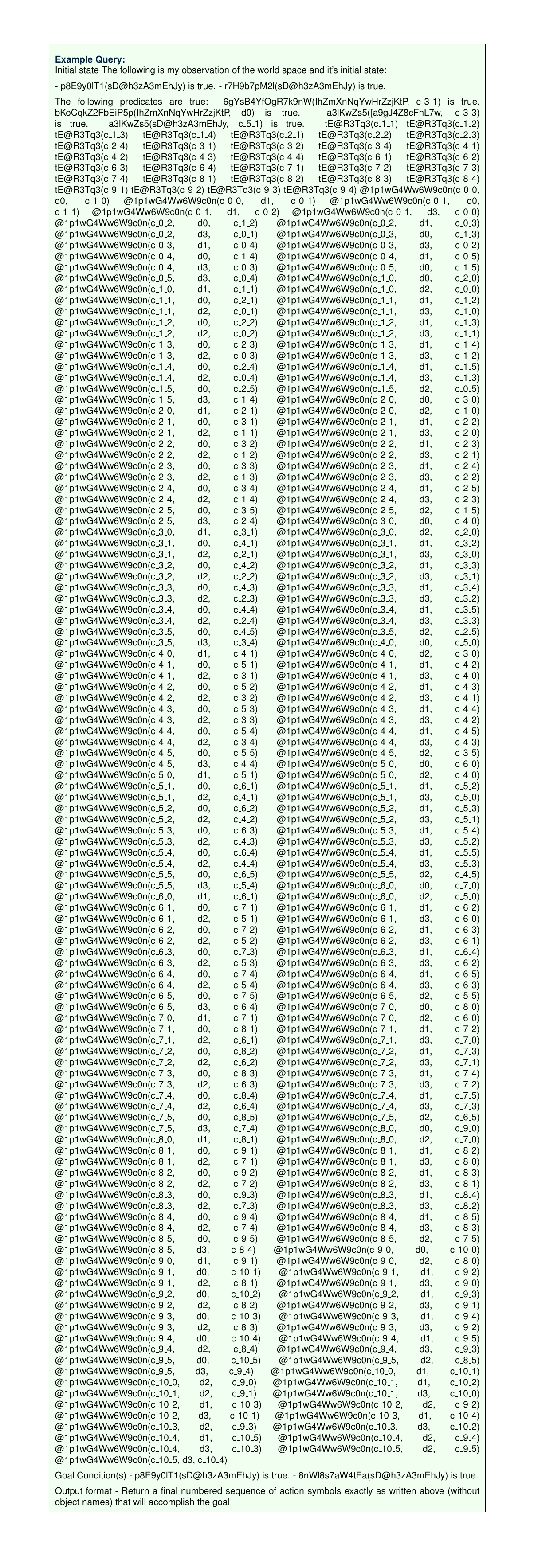}%
  \caption{Unlock query with an example of \texttt{Symbolic-Swapper} generated symbolic vocabulary.}
  \label{fig:unlock_grpo_query}
\end{figure*}

\begin{figure*}[!h]
  \centering
  \includegraphics[width=0.9\linewidth]{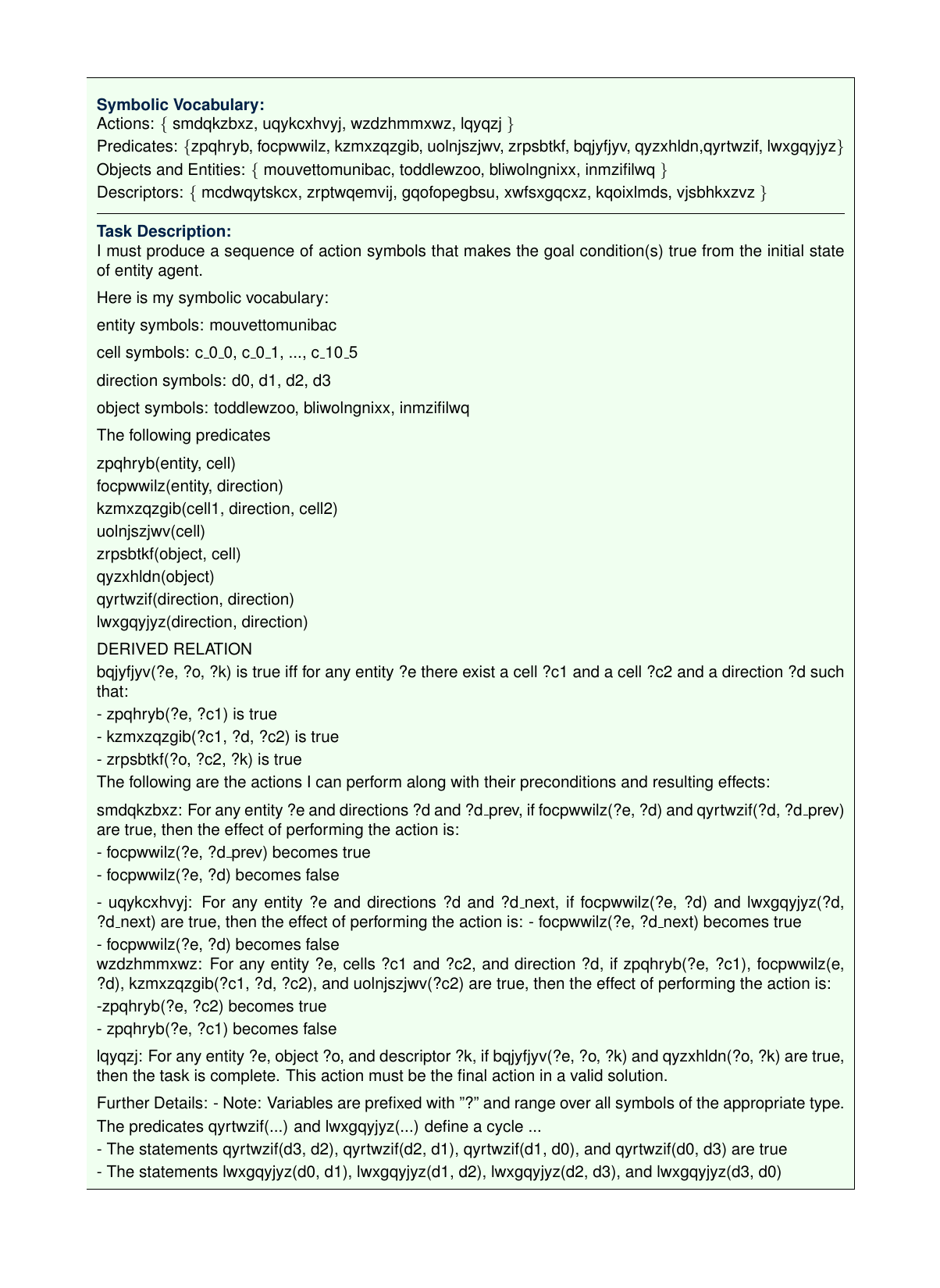}%
  \caption{Go To Object task description with an example of \texttt{Symbolic-Swapper} generated symbolic vocabulary.}
  \label{fig:gto_grpo}
\end{figure*}

\begin{figure*}[!h]
  \centering
  \includegraphics[width=0.65\linewidth]{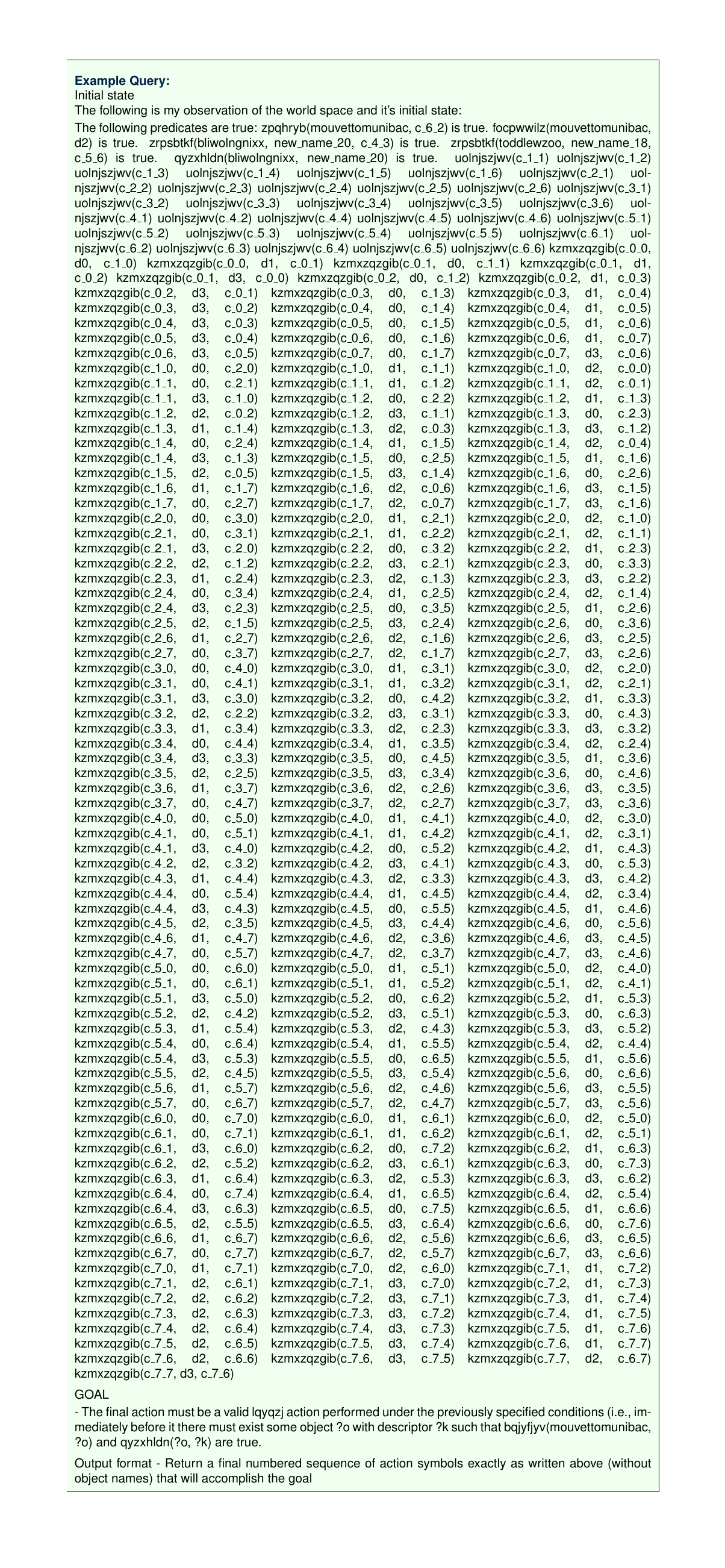}%
  \caption{Go To Object query with an example of \texttt{Symbolic-Swapper} generated symbolic vocabulary.}
  \label{fig:gto_grpo_query}
\end{figure*}

\FloatBarrier




\end{document}